\documentclass[lettersize,journal]{IEEEtran}
\usepackage{amsmath,amssymb,amsfonts}
\usepackage{mathtools}
\let\classAND\AND
\let\AND\relax
\usepackage{algorithmic}

\let\AND\classAND
\AtBeginEnvironment{algorithmic}{\let\AND\algoAND}

\let\classOR\OR
\let\OR\relax

\let\OR\classOR
\AtBeginEnvironment{algorithmic}{\let\OR\algoOR}

\usepackage{changes}
\usepackage{algorithm}
\usepackage{array}
\usepackage[caption=false,font=normalsize,labelfont=sf,textfont=sf]{subfig}
\usepackage{textcomp}
\usepackage{stfloats}
\usepackage{url}
\usepackage{verbatim}
\usepackage{graphicx}
\usepackage{cite}
\usepackage{bm,bbm}

\newcommand{\cX}{{\mathcal X}}
\newcommand{\cY}{{\mathcal Y}}
\newcommand{\cZ}{{\mathcal Z}}

\def\mc{\mathcal}

\def\mbf{\mathbf}
\def\msf{\mathsf}

\DeclarePairedDelimiter\parens{\lparen}{\rparen}  
\DeclarePairedDelimiter\abs{\lvert}{\rvert}
\DeclarePairedDelimiter\norm{\lVert}{\rVert}

\DeclarePairedDelimiter\braces{\lbrace}{\rbrace}
\DeclarePairedDelimiter\bracks{\lbrack}{\rbrack}

\newcommand{\R}{\mathbb{R}}

\newcommand{\defeq}{\triangleq}

\DeclareMathOperator{\sgn}{sgn}
\newcommand{\Indic}[1]{\mathbbm{1}\parens*{#1}} %

\DeclareFontFamily{U}{matha}{\hyphenchar\font45}
\DeclareFontShape{U}{matha}{m}{n}{
      <5> <6> <7> <8> <9> <10> gen * matha
      <10.95> matha10 <12> <14.4> <17.28> <20.74> <24.88> matha12
      }{}
\DeclareSymbolFont{matha}{U}{matha}{m}{n}
\DeclareFontSubstitution{U}{matha}{m}{n}
\DeclareFontFamily{U}{mathx}{\hyphenchar\font45}
\DeclareFontShape{U}{mathx}{m}{n}{
      <5> <6> <7> <8> <9> <10>
      <10.95> <12> <14.4> <17.28> <20.74> <24.88>
      mathx10
      }{}
\DeclareSymbolFont{mathx}{U}{mathx}{m}{n}
\DeclareFontSubstitution{U}{mathx}{m}{n}
\DeclareMathDelimiter{\vvvert}{0}{matha}{"7E}{mathx}{"17}
\DeclarePairedDelimiterX{\matnorm}[1]
  {\vvvert}
  {\vvvert}
  {\ifblank{#1}{\:\cdot\:}{#1}}

\DeclareMathOperator*{\argmin}{\mathrm{argmin}}

\DeclareMathOperator{\AND}{\mathsf{AND}}
\DeclareMathOperator{\OR}{\mathsf{OR}}

\NewDocumentCommand{\expect}{ e{_} s o >{\SplitArgument{1}{|}}m }{%
  \operatorname{\mathbb{E}}%
  \IfValueT{#1}{{\!}_{#1}}%
  \IfBooleanTF{#2}{%
    \expectarg*{\expectvar#4}%
  }{%
    \IfNoValueTF{#3}{%
      \expectarg{\expectvar#4}%
    }{%
      \expectarg[#3]{\expectvar#4}%
    }%
  }%
}
\NewDocumentCommand{\expectvar}{mm}{%
  #1\IfValueT{#2}{\nonscript\;\delimsize\vert\nonscript\;#2}%
}
\DeclarePairedDelimiterX{\expectarg}[1]{[}{]}{#1}
\newcommand{\E}{\ensuremath{\mathbb{E}}}
\NewDocumentCommand{\prob}{ e{_} s o >{\SplitArgument{1}{|}}m }{%
  \operatorname{\mathbb{P}}%
  \IfValueT{#1}{{\!}_{#1}}%
  \IfBooleanTF{#2}{%
    \probarg*{\probvar#4}%
  }{%
    \IfNoValueTF{#3}{%
      \probarg{\probvar#4}%
    }{%
      \probarg[#3]{\probvar#4}%
    }%
  }%
}
\NewDocumentCommand{\probvar}{mm}{%
  #1\IfValueT{#2}{\nonscript\;\delimsize\vert\nonscript\;#2}%
}
\DeclarePairedDelimiterX{\probarg}[1]{(}{)}{#1}
\renewcommand{\P}{\ensuremath{\mathbb{P}}}
\newcommand{\condon}{\,\ifnum\currentgrouptype=16 \middle\fi|\,} %

\newcommand{\Unif}{\mathsf{Unif}}

\newcommand{\datadist}{\pi}
\newcommand{\traindist}{\tau}
\newcommand{\param}{\theta}

\newcommand{\rparam}{\boldsymbol{\param}}
\newcommand{\rtest}{\mbf{x}}
\newcommand{\dtrain}{\mc{D}_{\mathrm{train}}}
\newcommand{\dtest}{\mc{D}_{\mathrm{test}}}
\newcommand{\dparam}{\mc{D}_{\mathrm{param}}}

\newcommand{\Ntrain}{N_{\mathrm{train}}}%
\newcommand{\Ntest}{N_{\mathrm{test}}}%

\newcommand{\Nmodels}{M}%
\newcommand{\Ncandmodels}{M'}%
\newcommand{\Nensemble}{M_{\mathrm{ens}}}
\newcommand{\mpos}{m^{+}}
\newcommand{\mneg}{m^{-}}
\newcommand{\gapclass}{\mc{M}}

\newcommand{\emprisk}{\hat{R}}
\newcommand{\testacc}{A}
\newcommand{\churn}{C}
\newcommand{\expectedcdf}{F_{\tau \times \pi}}
\newcommand{\condcdf}{\bar{F}_{\pi|\dparam}}

\newcommand{\avgcdf}{\hat{\bar{G}}}
\newcommand{\avgcdfinterp}{\hat{G^{l}}}
\newcommand{\cdfavg}{\hat{\bar{H}}'}
\newcommand{\indcdf}{\hat{G}_{k}}
\newcommand{\indtruecdf}{G_k}
\newcommand{\candtruecdf}{G_{0}}
\newcommand{\candcdf}{\hat{G}_{0}}
\newcommand{\multcandcdf}{\hat{G}_{l}}
\newcommand{\robustlevel}{\nu}
\newcommand{\testthresh}{\gamma}

\newcommand{\ensemblemodel}{\bar{m}}

\newcommand{\normifty}[1]{\norm*{#1}_{\infty}}

\newcommand{\bhtproxy}{\underset{\mc{\tilde{H}}_{0}}{\overset{\mc{\tilde{H}}_{1}}{\gtrless}}}
\newcommand{\randomness}{\msf{S}}

\newcommand{\trimmed}{\mc{R}}
\newcommand{\contam}{\mc{B}}
\newcommand{\distros}{\mc{P}}

\newcommand{\randinit}{\randomness_{\mathrm{init}}}
\newcommand{\randbatch}{\randomness_{\mathrm{batch}}}

\newcommand{\randall}{\randomness_{\mathrm{all}}}
\newtheorem{lemma}{Lemma}
\newtheorem{theorem}{Theorem}
\newtheorem{corollary}{Corollary}
\newtheorem{remark}{Remark}

\begin{document}

\title{Measuring training variability from stochastic optimization using robust nonparametric testing}

\author{Sinjini Banerjee~\IEEEmembership{Student Member,~IEEE}, %
Tim Marrinan, Reilly Cannon, Tony Chiang, Anand D.~Sarwate~\IEEEmembership{Senior Member,~IEEE}%
\thanks{Manuscript received June 28, 2024; revised March 24, 2025.}
\thanks{
This work was partially supported by the Statistical Inference Generates kNowledge for Artificial Learners (SIGNAL) program at Pacific Northwest National Laboratory (PNNL), as well as by the Mathematics for
Artificial Reasoning in Science (MARS) initiative via the Laboratory Directed Research and Development (LDRD) Program at PNNL. PNNL is a multi-program national laboratory operated for the U.S.~Department of Energy (DOE) by Battelle Memorial Institute under Contract No.~DE-AC05-76RL0-1830.}
\thanks{%
S. Banerjee and A. D. Sarwate are with the
    Department of Electrical and Computer Engineering, %
    Rutgers, The State University of New Jersey, %
    Piscataway, NJ 08854, USA. %
    Email: \texttt{\{sb1977, anand.sarwate\}@rutgers.edu}. %
T. Marrinan and R. Cannon are with the %
    Pacific Northwest National Lab, %
    Richland, WA 99352, USA. %
    Email: \texttt{\{timothy.marrinan,reilly.cannon\}@pnnl.gov}.
T. Chiang is affiliated with the 
    Pacific Northwest National Laboratory and 
    ARPA-H, %
    Bethesda, Maryland, 20892. %
    Email: \texttt{tony.chiang@arpa-h.gov}.
}
}

\maketitle

\begin{abstract}
Deep neural network training often involves stochastic optimization, meaning each run will produce a different model. This implies that hyperparameters of the training process, such as the random seed itself, can potentially have significant influence on the variability in the trained models. Measuring model quality by summary statistics, such as test accuracy, can obscure this dependence. %
We propose a robust hypothesis testing framework and a novel summary statistic, the $\alpha$-trimming level, to measure model similarity.
Applying hypothesis testing directly with the $\alpha$-trimming level is challenging because we cannot accurately describe the distribution under the null hypothesis. Our framework addresses this issue by determining how closely an approximate distribution resembles the expected distribution of a group of individually trained models and using this approximation as our reference.
We then use the $\alpha$-trimming level to suggest how many training runs should be sampled to ensure that an ensemble is a reliable representative of the true model performance. 
We also show how to use the $\alpha$-trimming level to measure model variability and demonstrate experimentally that it is more expressive than performance metrics like validation accuracy, churn, or expected calibration error when taken alone.
An application of fine-tuning over random seed in transfer learning illustrates the advantage of our new metric.
\end{abstract}


\section{Introduction}

Deep learning models have achieved state-of-the-art performance on complex tasks in healthcare, education, cyber-security, and other critical domains. Training these models takes significant time, energy, and hence financial resources. Training algorithms use stochastic optimization for non-convex objectives, meaning that models produced by different training runs, in general, converge to different solutions. It is clear that these trained models correspond to distinct functions, but is this a distinction without a difference? Models with a similar objective value and validation/test accuracy may still differ significantly.

In practice, models are often retrained as new data arrives. This necessitates algorithmic and architectural changes in state-of-the-art models to improve their performance on new data. The run-to-run variability in training models makes it difficult to conclude if a specific initialization or hyperparameter made a meaningful difference in model performance or if it just ``got lucky'' due to the presence of randomness in the optimization. Without this knowledge, comparing training configurations to assess relative quality becomes difficult. 

\IEEEpubidadjcol

We consider a stylized model of a machine learning training process. A practitioner trains $\Nmodels$ models in an identical manner, using fresh randomness for each training run, but they must output only a single model (due to computational constraints in deployment, perhaps). If test accuracy is the criterion they use, then they may have many models with nearly the same test accuracy. This work proposes a new framework to measure how representative a given model is of the training process.

\IEEEpubidadjcol

Gundersen et al.~\cite{gundersen2023sources} provided a taxonomy of barriers to reproducibility in machine learning practice: they identified randomness in model initialization, batch shuffling during mini-batch stochastic gradient descent (SGD), and data sampling as major sources of variability during training.
There are two sources of randomness in this framing of the problem: the sampling randomness of the test data, and the randomness from the stochastic optimization. The latter is under the control of the pracitioner.
The importance of using random seeds as part of hyperparameter tuning in deep neural network (DNN) training has been previously highlighted in the literature~\cite{henderson2019deep,reimers2017reporting,melis2017state,pmlr-v97-bouthillier19a,picard2021torch}: these works found the effect of random seed on model performance to be statistically significant. Other works have focused on understanding the random seed effect on specific sources of randomness in training. For example, Fort et al.~\cite{fort2020deep} found that initialization has a larger effect than batch order in SGD on model performance, while Bouthillier et al.~\cite{bouthillier2021accounting} showed the opposite. Summers and Dinneen~\cite{summers2021nondeterminism}, and Jordan~\cite{jordan2023calibrated} show that networks converge to vastly different values even with the change of a single bit of network parameters during initialization. While studying run-to-run variability in pre-trained BERT models over different random seeds, Dodge et al.~\cite{dodge2020finetuning} noted a validation performance gain of 7\% over previously reported results, highlighting the importance of fine-tuning over the random seed. All the above-mentioned works have assessed the impact randomness has on the \emph{validation accuracy}, or \emph{churn}~\cite{NIPS2016_dc5c768b} between models, which quantifies how two models differ in their predictions on the same test point. These summary statistics only focus on the \emph{decisions} made by predictive (classification) models and do not directly assess differences in the functions learned by these models.

\begin{figure}[!t]
\centering
\includegraphics[width=1.4in]{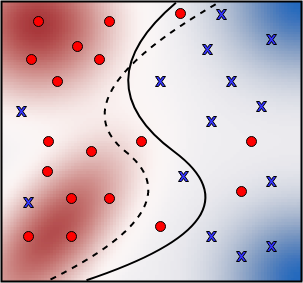}
\caption{Hypothetical decision boundaries corresponding to two models with the same accuracy. The shaded regions represent the underlying data distribution.}
\label{fig:measures_example}
\end{figure}

In this paper, we assess DNN training variability over random seeds using the \emph{network outputs used to make the decision}. Figure \ref{fig:measures_example} illustrates the difference: the solid and dashed lines represent two decision boundaries between red (circle) and blue (cross). Accuracy measures incorrect decisions, and the churn is determined by the region between the two curves. 
If we think of the training algorithm as generating a random sample from a function space, we can use other tools to understand model variability. As a first step, we can examine the distribution of the pre-thresholded outputs (the \emph{logit gap}) from functions learned by different runs of a fixed DNN architecture. To that end, we adopt a non-parametric hypothesis testing framework to measure the similarity of functions learned by DNN models that vary in random seeds.

\section{Problem setup}\label{sec: Problem Setup}

\noindent \textbf{Notation:} For a positive integer $n$, let  $[n] = \{1,2,\ldots,n\}$. Let $\Indic{.}$ denote the indicator function, such that $\Indic{x \le t}=1$ if $x \le t$, and $0$ otherwise. Random variables will be denoted in boldface, with realizations being non-bolded, so $\rparam$ is a random variable and $\param$ is a realization. We consider training a DNN for predicting an output taking values in a set $\cZ$ with input from a (feature vector) $x$ taking values in a space $\cX$. A DNN with a given architecture is specified by a set of parameters (e.g.~the weights) $\param$ taking values in parameter space, $\varTheta$.

\subsection{The learning task}

We will restrict attention to \emph{binary classification} using a DNN with parameters $\param$. Given a label space $\cY$, a training algorithm takes a training set, $\dtrain = \{ (x_i, y_i) \in \cX \times \cY \colon i \in [\Ntrain] \}$, estimates the weights $\param$ that (approximately) minimize the empirical loss $\emprisk(\param; \dtrain)$ over the training data, and assigns a label $y \in \cY$ using a \emph{softmax} operation. 
 We model this by assuming the DNN takes an input $x$ and computes functions $\mpos(x \mid \param)$ and $\mneg(x \mid \param)$, with the predicted label  $\hat{y}(x; \param)$ being $+1$ if $\mpos(x \mid \param) \ge \mneg(x \mid \param)$ and $-1$ otherwise. A Bayesian interpretation of this rule assumes the data is generated according to an (unknown) distribution $\datadist(x,y)$, with a likelihood function $\bar{\datadist}(x | y)$, and a uniform prior $\Unif[\cY]$. The functions $\mpos(x \mid \param)$ and $\mneg(x \mid \param)$ for the positive and negative classes can be converted into posterior probability estimates:
    \begin{align}
    \hat{\datadist}(y = 1 \mid x) &= \frac{\exp(\mpos(x \mid \param))}{ \exp(\mpos(x \mid \param)) + \exp(\mneg(x \mid \param))}. \\
    \hat{\datadist}(y = -1 \mid x) &= \frac{\exp(\mneg(x \mid \param))}{ \exp(\mpos(x \mid \param)) + \exp(\mneg(x \mid \param))}.
    \end{align}
The prediction function is then the maximum \textit{a posteriori} (MAP) estimate:
    \begin{align}
    \hat{y}(x; \param) &= \sgn\parens*{ \hat{\datadist}(y = 1 \mid x) - \hat{\datadist}(y = -1 \mid x)} \notag\\
    &= \sgn\parens*{\mpos(x \mid \param) - \mneg(x \mid \param)}.
    \end{align}
We can then assume the learned function is $m(x; \param) =  \mpos(x \mid \param) - \mneg(x \mid \param)$ which is an approximation to the (unknown) log-likelihood ratio $\log\frac{\bar{\datadist}(x | y = 1)}{ \bar{\datadist}(x | y = -1)}$. We therefore refer to $\mpos(x \mid \param)$ and $\mneg(x \mid \param)$ as \emph{logits} and $m(x ; \param)$ as the \emph{logit gap}.
In our setting, for a given $\param \in \varTheta$, the network computes the function $m(x;\param)$ (logit gap function), and the DNN defines a family of functions $\gapclass = \{m(x;\param): \cX \rightarrow \cZ \colon \param \in \varTheta \}$. The goal of \emph{training} a DNN is to find a ``good'' setting for the parameters $\param$ or, equivalently, to find a ``good'' function $m \in \gapclass$. 

Almost all DNN training algorithms use \emph{stochastic optimization}, making them approximate in two ways. First, they will generally converge to a local minimum because the risk minimization problem is non-convex. Second, randomization means the estimated parameters $\rparam$ are random variables. Two runs of the same training algorithm on the same training set $\dtrain$ can produce different functions. One natural question is to ask how different these functions are. We can try to answer this using a \emph{test set}, $\dtest = \{ (x_j, y_j) \colon j \in [\Ntest] \}$. If we assume $\dtest$ is drawn i.i.d. from the data distribution $\datadist$, then given a trained model $m(x; \param)$, the set of values $\{ m(x_j ; \param) \colon j \in [ \Ntest]\}$ is also an i.i.d.~sample from a distribution on $\R$ induced by $\datadist$.

\subsection{Trained models and reference functions}\label{subsec: true cdfs}

\begin{figure}[tbh]
\centering
\includegraphics[width=0.489\textwidth]{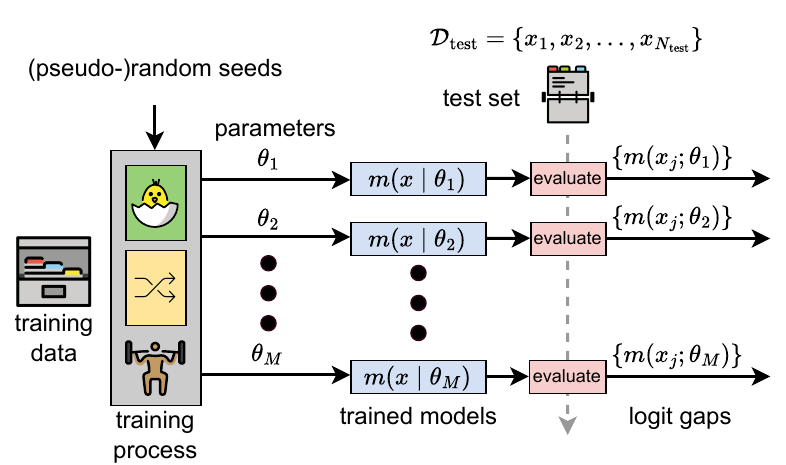}
\caption{The experiment design. Randomness is used within the training algorithm for initialization and batch selection. We train (or fine-tune) $\Nmodels$ models independently using the same training data. Each model is then evaluated on the test set. The resulting values are used to form empirical CDFs.}
\label{fig:experiment}
\end{figure}

We consider the experiment shown in Figure \ref{fig:experiment}: we train $\Nmodels$ models independently by varying the random seed and using the same training set. We then evaluate the corresponding models on the test set.
Let $\traindist$ be the distribution of $\rparam$ corresponding to the randomness in the training algorithm run on data set $\dtrain$.
Then $\dparam = \{\param_1,\param_2,\ldots, \param_M\}$ is sampled i.i.d.~$\sim \traindist$. 
Let $\distros$ be the set of cumulative distribution functions (CDFs) on $\R$.
Given a fixed DNN architecture, the training set $\dtrain$, and a stochastic training algorithm, %
the $M$ i.i.d. samples of parameters induce $M$ i.i.d.  samples of functions 
$\{ m(x; \param_k) : k \in [\Nmodels]\},$ taking values in $\gapclass$. 

Given a realization of a parameter $\param_k$ and $\rtest \sim \datadist$, we define the CDF of the model $m(\rtest; \param_k)$ by 
\begin{align}\label{eq:singlemodelCDF}
     \indtruecdf(t) = \E_{\datadist|\param_{k}} \bracks*{\Indic{ m(\rtest; \param_{k}) \le t }}
\end{align}
Now, suppose $\rparam \sim \traindist$ and $\rtest \sim \datadist$ are drawn independently. The expected CDF over both is
     \begin{align}
     \expectedcdf(t) &= \E_{\traindist \times \datadist}\bracks*{ \Indic{m(\rtest; \rparam) \le t}}.
         \label{eq:expectedCDF}
     \end{align}
    
Since we do not know $\traindist \times \datadist$, we can consider an approximate empirical version of $\expectedcdf$ by conditioning \eqref{eq:expectedCDF} on the sampled parameters $\dparam$. The result is the following function:
\begin{align}\label{eq:expectedCDF params}
   \condcdf(t) 
   &= \E_{\datadist | \dparam}\bracks*{ 
    \frac{1}{\Nmodels} \sum_{k=1}^{\Nmodels}
        \Indic{m(\rtest; \param_{k}) \le t}}
    \notag\\
   &= \frac{1}{\Nmodels} \sum_{k=1}^{\Nmodels}
    \E_{\datadist | \dparam}\bracks*{
        \Indic{m(\rtest; \param_{k}) \le t}}, 
    \notag \\
   &= \frac{1}{\Nmodels} \sum_{k=1}^{\Nmodels}
    \E_{\datadist | \param_k}\bracks*{
        \Indic{m(\rtest; \param_{k}) \le t}},
\end{align}
by the linearity of expectation. Note that the terms inside the sum are the CDFs in \eqref{eq:singlemodelCDF}.

\subsection{Empirical CDFs, reference functions, and ensembles}\label{subsec: empirical cdfs}

We can only access data distribution $\datadist$ through samples from $\dtest$. In our subsequent hypothesis tests, we will use empirical cumulative distribution functions (eCDFs) in the test statistics. Given $N$ samples from $\datadist$ we can compute the eCDF for a model with parameter $\param_k$
\begin{align}\label{eq:Individual ecdf}
     \indcdf(t) = \frac{1}{N} \sum_{j=1}^{N} \Indic{ m(x_j; \param_{k}) \le t }.
\end{align}
We can average the eCDFs from $\Nmodels$ models to form what we call the \emph{reference function}
    \begin{equation}\label{eq:avg_cdf}
    \avgcdf(t)=\frac{1}{\Nmodels}\sum_{k=1}^{\Nmodels} 
        \parens*{ 
        \frac{1}{ N}\sum_{j=1}^{ N}  
            \Indic{m(x_{j},\param_{k}) \le t } 
        }.
    \end{equation}

In our framework, we will also consider \emph{ensemble models} in which we average the logit gaps. If we take a subset $(\param'_1, \param'_2, \ldots, \param'_{\Nensemble})$ of $\Nensemble$ models from our $\Nmodels$ models, the corresponding ensemble is
    \begin{align}
    \ensemblemodel(x_{j})=\frac{1}{\Nensemble} \sum_{k=1}^{\Nensemble} m(x_{j},\param'_{k}).
    \end{align}
The eCDF of this ensemble model is
\begin{equation}\label{eq:ensemble cdf}
    \cdfavg(t)= \frac{1}{ N} \sum_{j=1}^{ N}
            \Indic{ \ensemblemodel(x_{j})  \leq t  }. 
\end{equation}

We are interested in testing whether a new model with parameter $\param_0$ is close to the expected model from the training process. We formulate this using hypothesis tests that compare $\candtruecdf$ corresponding to $\param_0$ as defined in \eqref{eq:singlemodelCDF} with the average in $\condcdf$. 
By switching on or off different uses of randomization in training, we induce different distributions $\traindist$ on the parameters. For example,
we can use deterministic initialization or fixed batch ordering. Under these scenarios, we can generate $\Nmodels$ models and analyze the variability of these models using the reference function \eqref{eq:avg_cdf}, which captures the consensus of the collection of trained models.

\begin{remark} %
While we have described the problem in this section for binary classifiers and the logit gap, note that this formulation can be used for any machine learning model by taking a single scalar measurement function $m \colon \mc{X} \times \varTheta \to \R$ applied to $\dtest$. For binary classifiers, the logit gap is a natural choice and makes the comparison to the validation accuracy more interpretable.
\end{remark}

\section{Robust and non-parametric testing}\label{KS Test}

Ideally, we want to estimate the variability of trained models by measuring the discrepancy between a candidate model, $\candtruecdf$, and the expected CDF $\expectedcdf$ defined in \eqref{eq:expectedCDF}. Without access to $\traindist \times \datadist$, we instead consider the CDF, $\condcdf,$ averaged over $\dparam$ as defined in  \eqref{eq:expectedCDF params}. We formulate this as a a one-sided hypothesis test: the null hypothesis is that $\candtruecdf$ is the same as $\condcdf$.

\subsection{Classical one- and two-sample KS tests}

We can use non-parametric hypothesis testing to formulate the comparison between $\candtruecdf$ and $\condcdf$. Given $N$ samples, $\{x_j : j \in [N]\} \sim \datadist$, we can compute the logit gaps, $\{ m(x_j; \param_0) \colon j \in [ N] \}$. The null hypothesis for the one-sided test is
    \begin{align}
    \mc{H}_0^{\mathrm{KS1}} &\colon \{ m(x_j; \param_0) \colon j \in [N] \} \sim \condcdf.
    \end{align}
That is, is the candidate, $\candtruecdf,$ the same as $\condcdf$? The classical Kolmogorov-Smirnov (KS) test uses the eCDF, $\candcdf,$ defined in \eqref{eq:Individual ecdf} to compute the test statistic, $\norm{ \condcdf - \candcdf}_{\infty}$.

The KS test statistic cannot be evaluated without a closed-form expression for $\condcdf$. 
However, we can still employ a two-sample KS test, which leads us to the null hypothesis
    \begin{align}
    \mc{H}_0^{\mathrm{KS2}} &\colon \candtruecdf = \condcdf.
    \end{align}
Unfortunately, this hypothesis also relies on explicit knowledge of $\condcdf$. To get around this, we propose using $\avgcdf$ as a proxy for $\condcdf$.    
In this case we draw $2 N$ samples $\{x_j : j \in [2 N]\} \sim \datadist$ and use half to compute the reference function $\avgcdf$ in \eqref{eq:avg_cdf} and half to compute $\candcdf$ and use the test statistic $\norm{\avgcdf - \candcdf}_{\infty}$. The following result shows that $\avgcdf$ is a reasonable proxy for $\condcdf$.

\begin{theorem}\label{Main theorem}
Let $\avgcdf$ and $\condcdf$ be given by  \eqref{eq:avg_cdf} and \eqref{eq:expectedCDF params}. Then for any $\delta_{b} > 0$, 
    \begin{align}
    \label{eq:dkw:ref}
    \P_{\datadist|\dparam}\left( \normifty{\condcdf- \avgcdf} > \delta_{b} \right) \le \epsilon_{b},
    \end{align}
where $\epsilon_{b} = 2\Nmodels \exp(-2  N \delta_{b}^{2})$.
\end{theorem} 

The proof is in Section \ref{sec: Analysis} of the Appendix and follows from the Dvoretzky-Kiefer-Wolfowitz (DKW) inequality~\cite{dvoretzky1956asymptotic} and a union bound over $\{\param_j\}$.

We can use the two-sample DKW inequality~\cite{WeiD:12dkw} to set a threshold for the test. The DKW inequality shows that 
 \begin{align}
  \prob_{\datadist|\dparam}*{ 
    \label{eq:dkw:cand}
     \normifty{\candcdf - \avgcdf}  > \delta_{a}
    }
    \leq \epsilon_{a},
    \end{align}
where $\epsilon_{a} = C \exp({- N \delta_{a}^2)}$. %
Given a target $\epsilon_{a}$ we can set the threshold for the test as
    \begin{align}\label{eq: Two samp threshold}
    \delta_{a} = \sqrt{\frac{1}{N} \ln{\frac{C}{\epsilon_{a}}}}.
    \end{align}
Here $C = e$ for general $N$ and for $N \ge 458$ we can take $C = 2$~\cite[Theorem 1]{WeiD:12dkw}.  
Using the the triangle inequality, we have the following corollary.

\begin{corollary}
Under the assumptions of Theorem \ref{Main theorem}, for any $\delta_a > 0$, $\delta_b > 0$ and sample size $N \ge 458$, we have
\begin{align}\label{eq: true hypothesis test bound}
    &\prob_{\datadist|\dparam}*{
        \normifty{\condcdf - \candcdf} \le \delta_{a} + \delta_{b}
        } \notag \\
    & \qquad \qquad \ge 1 - 2 M \exp\left( - 2 N \delta_b^2 \right) - 2 \exp( - N \delta_a^2 ).
\end{align}
\end{corollary}

Provided we have enough samples, we can set thresholds for this test to achieve a guaranteed false alarm (type I error) probability. 
However, in large sample settings, the KS-test often rejects the null because even small changes in the sample can result in a significant shift in the $L_{\infty}$-norm~\cite[p.245]{gibbons1992nonparametric}.

\subsection{Robust statistics and trimming}\label{sec: Robust KS test}

Modern machine learning models are complex and often have many more parameters than training points. Even if we use a very large number of models, $\Nmodels$, to compute $\avgcdf$ and estimate $\condcdf$, we may not have a good approximation to the expected CDF, $\expectedcdf$. %
We want to represent this epistemic uncertainty and alleviate the over-sensitivity of the KS test in a more structured way. We can capture some of this uncertainty in the null hypothesis by replacing the null hypothesis with a composite hypothesis: this is the approach taken in~\emph{robust statistics}~\cite{HuberR:09robustbook}.

We describe the approach for general distributions and later specialize it to our setting. Consider a given distribution $P_0$ and define a new null hypothesis as a set of distributions close to $P_0$. Given a metric or divergence $d(\cdot,\cdot)$, radius $\alpha$, and distribution $P_0$, define the $d$-ball of radius $\alpha$ by $\contam_d(P_0,\alpha) = \{ P \in \distros : d( P_0, P) \le \alpha\}$.
While we could take $d(\cdot,\cdot)$ to be any metric or divergence between probability distributions, the classical approach in robust statistics~\cite{HuberR:09robustbook} uses an $L_1$ ball corresponding to a \emph{contamination neighborhood}
    \begin{align}
    \contam_1(P_{0},\alpha) = \{ P : P = (1 - \alpha) P_0 + \alpha Q,\ Q \in \distros \}.
    \end{align}
The interpretation is that up to an $\alpha$-fraction of samples may come from an unknown ``outlier'' distribution $Q$.

In this paper, we use the $L_1$ ball not only because it is the default in robust statistics but also because of the connection to \emph{impartial trimming}~\cite{GORDALIZA1991162}.
Given a distribution $P_1 \in \distros$ and a scalar $\alpha \in [0,1]$, the set of $\alpha$-trimmings of $P_1$ is defined by,
    \begin{align} \label{eq:def:trimming}
    \trimmed_{\alpha}(P_1) =  \braces*{ P \in \distros \colon P \ll P_1,\ \frac{dP}{dP_1} \leq \frac{P_1}{(1 - \alpha)}~\text{a.s.}},
    \end{align}
where $P \ll P_1$ denotes that $P$ is absolutely continuous with respect to $P_1$~\cite{alvarez2008trimmed}. Trimmings are related to contamination neighborhoods~\cite{alvarez2011uniqueness}:
\begin{align}
    P_1 \in \contam_1(P_{0},\alpha) \Leftrightarrow P_{0} \in  \trimmed_{\alpha}(P_1),
    \label{eq:contam:trim}
\end{align}

A recent work by del Barrio, Inouzhe, and Matr\'{a}n~\cite{del2020approximate} formulated a robust hypothesis test using the connection between contamination and trimming. Given a sample generated from an unknown distribution $P$, their null hypothesis\footnote{The superscript is the initials of del Barrio, Inouzhe, and Matr\'{a}n~\cite{del2020approximate}.} is:
    \begin{align}
    \mc{H}_0^{\mathrm{BIM}} \colon P \in \contam_1(P_0,\alpha), 
    \label{eq:BIM:contamH0}
    \end{align}
where $P_0$ is a known distribution. Because of the connection to $\alpha$-trimming in \eqref{eq:contam:trim}, this hypothesis is equivalent to stating $P_0 \in \trimmed_{\alpha}(P)$. They, therefore, take the following as their null hypothesis:
    \begin{align}
    \mc{H}_0^{\mathrm{BIM}} \colon \inf_{\tilde{P} \in \trimmed_{\alpha}(P)} \norm{P_0 - \tilde{P}}_{\infty} = 0.    
    \label{eq:BIM:trimH0}
    \end{align}
Given a sample of $N$ points and its empirical eCDF $\hat{P}_{N}$, they propose the test statistic:
    \begin{align}
    T(\hat{P}_N) 
    = \inf_{\tilde{P} \in \trimmed_{\alpha}(\hat{P}_N)} \norm{P_0 - \tilde{P}}_{\infty}.
    \label{eq:delbarrio:teststat}
    \end{align}
This test statistic involves finding the closest $L_{\infty}$ approximation to $P_0$ in the set of $\alpha$-trimmings of the eCDF $\hat{P}_N$. This change is important because test statistics using the contamination-based null in \eqref{eq:BIM:contamH0} could entail an optimization over the $L_1$ ball $\contam_1(P_0,\alpha)$, which would be infinite dimensional problem, wheras optimization over the set of $\alpha$-trimmings in \eqref{eq:delbarrio:teststat} is a finite dimensional problem.

Computing the optimizer of \eqref{eq:delbarrio:teststat} in the set of trimmings of $\hat{P}_N$ involves finding a reweighting of the samples such that downplaying the importance of a small fraction of contaminated samples (from the test set) would allow a KS-test to not reject the null hypothesis. We point the readers to Section \ref{subsec: Impartial trimming}, and \ref{subsec: Trimmed Kolmogorov-Smirnov distance} in the Appendix for a detailed description. 

\subsection{A new robust two-sample test}\label{subsec: a new robust two sample test}

To apply the trimming-based approach in our problem, we could take $P_0 = \condcdf$ and test using the eCDF $\hat{P}_N = \candcdf$ computed from $N$ samples. Because $\condcdf$ is not known, we cannot compute the test statistic in \eqref{eq:delbarrio:teststat}. In what follows, we assume that the support $S$ of the distributions is bounded\footnote{While logit gaps can in general take any values in $(-\infty,\infty)$, in practice, logit gaps of well-trained models are not too large. Large logit gaps would indicate a poorly calibrated model or the model being overconfident in one class. In our experiments we clip the logit gaps.} with length $|S|$: this assumption will let us set the threshold in our test.

\begin{figure}[t!]
 \centering
\centerline{\includegraphics[width=2.4in]{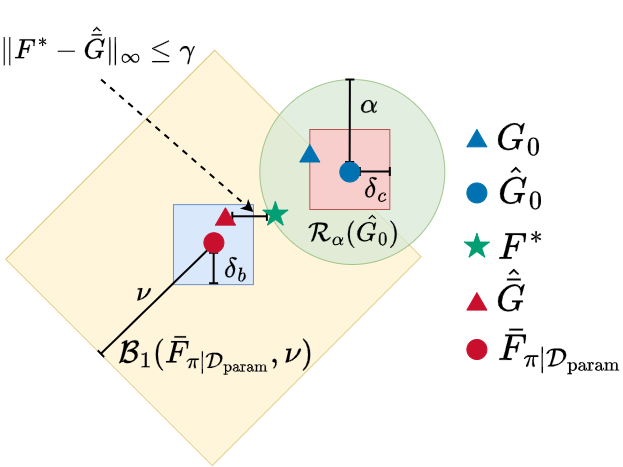}}
 \caption{Illustration of the parameters in the trimming-based two-sample test. The null hypothesis uses the $L_{1}$-contamination ball, $\contam_1( \condcdf,\robustlevel)$. The test will accept if the $L_{\infty}$ distance between $\condcdf$ and the set of $\alpha$-trimmings of $\candcdf$, $\trimmed_{\alpha}(\candcdf)$, is small. }
 \label{fig:robust:2sample}
\end{figure}

We propose using the following null hypothesis in a two-sample test:
    \begin{align}\label{eq: Robust null hypothesis}
    \mc{H}_0 \colon \candtruecdf \in \contam_1( \condcdf, \robustlevel).
    \end{align}
As before, we need to find a test statistic and threshold using $2 N$ samples from $\datadist$. We use $N$ samples to compute $\avgcdf$ and $N$ samples to compute $\candcdf$. The DKW inequality shows that for any $\delta_c > 0$,
    \begin{align}
        \norm{\candtruecdf - \candcdf}_{\infty} \le \delta_{c}
        \label{eq:DKW:cand}
    \end{align}
with probability $1 - 2 e^{-2 N {\delta}^2_c}$.

From the bounds in \eqref{eq:DKW:cand} and \eqref{eq:dkw:ref} and our assumption on the support, we have the following bounds:
    \begin{align} 
    \norm{\candtruecdf - \candcdf}_1 &\le |S| \delta_{c} 
    \label{eq:DKW1:cand}
    \\ 
    \norm{\condcdf - \avgcdf}_1 &\le |S| \delta_{b}
    \label{eq:DKW1:ref}
    \end{align}
with probability $1 - 2 e^{-2 N {\delta}^{2}_c}$ and $1 - 2 M e^{-2 N {\delta}_b^2},$ respectively.
The decision rule we propose is to accept the null if
    \begin{align}
    \min_{\tilde{F} \in \trimmed_{\alpha}(\candcdf)} \norm{ \tilde{F} - \avgcdf }_{\infty} \le \testthresh.
    \label{eq:rule:new}
    \end{align}
This decision rule finds the best $L_{\infty}$ approximation of the reference $\avgcdf$ in the $\alpha$-trimmings of the candidate eCDF $\candcdf$.

We want to understand the relationship between the contamination level $\robustlevel$ of the null, the trimming level $\alpha$, and the threshold $\testthresh$. Let $F^*$ be the minimizer of \eqref{eq:rule:new}:
    \begin{align}
    F^* = \argmin_{\tilde{F} \in \trimmed_{\alpha}(\candcdf)} \norm{ \tilde{F} - \avgcdf }_{\infty}.
    \end{align}
Suppose the null is accepted by~\eqref{eq:rule:new} so that $\norm{ F^* - \avgcdf }_{\infty} \leq \testthresh$. This implies
    \begin{align}
    \norm{ F^* - \avgcdf }_{1} \le |S| \testthresh.
    \end{align}
By \eqref{eq:contam:trim}, $F^* \in \trimmed_{\alpha}(\candcdf)$ means $\candcdf \in \contam_1(F^*, \alpha)$, so
    \begin{align}
    \norm{ F^* - \candcdf }_{1} \le \alpha.
    \end{align}
Thus by the triangle inequality,
    \begin{align}
    \norm{ \candcdf - \avgcdf }_{1} \le \alpha + |S| \testthresh.
    \end{align}
Combining this with the two high probability inequalities from \eqref{eq:DKW1:cand} and \eqref{eq:DKW1:ref} we have
    \begin{align}\label{eq: FinalL1bound}
    \norm{ \condcdf - \candtruecdf }_{1} \le \alpha + |S| ( \testthresh + \delta_b + \delta_c ).
    \end{align}
with probability $1 - \varepsilon$, where
    \begin{align}\label{eq: probFinalL1bound}
    \varepsilon \le  2 e^{-2 N {\delta}^{2}_{c}} + 2 \Nmodels e^{-2 N \delta_b^2}.
    \end{align}
For a fixed threshold $\testthresh$, we can estimate the trimming level, ${\alpha}$, needed for the $L_{\infty}$ distance in \eqref{eq:rule:new} to be less than or equal to the threshold. Consequently, for a fixed test threshold $\testthresh$, and with enough samples to make $\delta_{b}$ and $\delta_{c}$ small, we can estimate the contamination level $\robustlevel$, in the contamination neighborhood $\contam_1( \condcdf, \robustlevel)$ of $\condcdf$, for which the robust two-sample hypothesis test would fail to reject the null with high probability $\epsilon$. We use this in the next section to propose a new measure for analyzing model variability.
\section{Metrics to analyze model variability}\label{Metrics}

One of our goals is to use the framework from the robust two-sample test to design a measure of model discrepancy/variability. As we show in the experiments, this new measure can be more informative than existing measures.

\subsection{A new trimming-based metric}
 
In this section, we introduce how our trimming-based two-sample test can be used to define a new measure of discrepancy between a model with parameter $\param_0$ and the reference model formed using $\param_1, \param_2, \ldots, \param_{\Nmodels}$. In Algorithm \ref{alg:alpha_search}, we estimate the smallest $\alpha$ such that the test in \eqref{eq:rule:new} accepts. We do this by running the test $B$ times using bootstrap resampling from $\dtest$. In each resampling we increase the trimming level $\alpha$ until the test accepts. This is equivalent to finding a contamination level $\robustlevel$ for the null such that the test accepts. The output $\hat{\alpha}$ is the average over the $\alpha$ values from each resampling and is our proposed measure of discrepancy between model $\candcdf$ and $\avgcdf$. 

We compute $\avgcdf$ and $\candcdf$ by sampling uniformly from $\dtest$ of size $\Ntest = 2N$. %
Let $\mc{Z} =\{z_{j}\}_{j=1}^{2N}$ be the combined ordered arrangement of $2 N$ logit values from $\avgcdf$ and $\candcdf$.  For a fixed $\alpha$, we calculate the $\alpha$-trimming of $\candcdf$ that minimizes the $L_{\infty}$-distance to $\avgcdf$. We then check to see if this trimming is within the $L_{\infty}$ ball around $\avgcdf$ of radius $\testthresh$:
\begin{align}
\max_{z \in \mc{Z}} \min_{\tilde{F} \in \trimmed_{\alpha}(\candcdf)} \abs*{ \tilde{F}(z) - \avgcdf(z) } \le \testthresh,
\label{eq: robust2sample_ht}
\end{align}
We set our threshold to be $\testthresh = \delta_{a}$ by fixing the %
probability $\epsilon_{a}$ in \eqref{eq: Two samp threshold}, which follows from the two-sample DKW inequality introduced in $\eqref{eq:dkw:cand}$.

A small $\hat{\alpha}$ for a candidate model $\candtruecdf$ implies the level of trimming needed for it to be not rejected, meaning that it is close to $\avgcdf$, which in turn is close to $\candcdf$. In Section \ref{sec: experiments}, we compare $\hat{\alpha}$ to metrics like accuracy and churn that are commonly used to analyze model variability. We generate multiple trained models from the randomization in the training process. We can look at each trained model as a potential candidate and use the rest to compute a reference.

\begin{algorithm}
\caption{ Estimate $\hat{\alpha}$ measure}
\label{alg:alpha_search}
\begin{algorithmic}
\renewcommand{\algorithmicrequire}{\textbf{Input:}}
\renewcommand{\algorithmicensure}{\textbf{Output:}}
\REQUIRE Test set $\dtest$, trained models $\{ m_k \colon k \in [\Nmodels] \}$, candidate model $m_{0}$,  threshold $\delta_{a}$, trimming levels $\{\alpha_t\}_{t=1}^{T}$, bootstrap sampling number $B$.
\ENSURE trimming level estimate $\hat{\alpha}$.
\FOR {$b = 1$ to $B$}
\STATE Compute $\avgcdf$ in \eqref{eq:avg_cdf} using $\{ m_k \colon k \in [\Nmodels] \}$, and $\candcdf$ in \eqref{eq:Individual ecdf} using $m_0$ from two sets of ${N}$ samples resampled from $\dtest$.
\STATE Set $\mc{Z}=\{z_{j}\}_{j=1}^{2N}$ be the ordered set of logit values from $\avgcdf$ and $\candcdf$.
\STATE  $\msf{Reject} \gets 1$, $\hat{\alpha} \gets 0$, $t \gets 1$
\WHILE {$\msf{Reject} = 1$ and $t \le T$}
\STATE $\alpha \gets \alpha_{t}$
\STATE Use $\mc{Z}$ to compute \eqref{eq: robust2sample_ht}
\IF {Test \eqref{eq: robust2sample_ht} accepts}
\STATE $\msf{Reject} \gets 0$,
\STATE $\hat{\alpha}_b \gets \alpha$
\ELSE
\STATE $t = t + 1$
\ENDIF
\ENDWHILE
\ENDFOR\\
$\hat{\alpha} \gets \frac{1}{B} \sum_{b=1}^{B} \hat{\alpha}_b$
\end{algorithmic}
\end{algorithm}

\subsection{Other metrics for model variability}\label{sec: other metrics}
 
The \emph{accuracy} of a model is
    \begin{align}
    \testacc(\param) = \frac{1}{ N} 
        \sum_{j=1}^{ N} 
            \Indic{ \hat{y}(x_j; \param) = y_j }.
    \end{align}
The \emph{churn} is defined by
    \begin{align}\label{eq:churn}
    \churn(\param_1, \param_2) = 
        \sum_{j=1}^{ N} 
            \Indic{ \hat{y}(x_j; \param_1) \ne \hat{y}(x_j; \param_2) },
    \end{align}
which is the number of test points where the models disagree. Both accuracy and churn focus on the \emph{predictions} made by models and do not use information about the logit gap function $m(x;\param_{k})$ beyond its sign. Looking at $m(x;\param_{k})$ directly gives us other approaches to assess whether models are similar or not: two models may have similar accuracy and low churn but can have very different logit gaps. 

We also consider a third metric for the qualitative assessment of these models, the Expected Calibration Error (ECE)~\cite{naeini2015obtaining}. The ECE is a summary statistic of model calibration which measures the difference in accuracy and expected confidence and is obtained by partitioning the predictions into $R$ equally spaced bins $B_{r}$, 
\begin{align}\label{eq:ECE}
    \text{ECE}(\param) = \sum_{r=1}^{R} \frac{\abs*{B_{r}}}{ N} \abs*{ \testacc(B_{r};\param) - \text{CONF}(B_{r};\param)},
\end{align}
where  
    \begin{align}
    \testacc(B_{r};\param) &= \frac{1}{\abs*{B_{r}}}
        \sum_{j =1}^{\abs*{B_{r}}}
            \Indic{ \hat{y}(x_j; \param) = y_j } \\
    \text{CONF}(B_{r};\param) &= \frac{1}{\abs*{B_{r}}} \sum_{j =1}^{\abs*{B_{r}}}  \hat{\datadist}(y_j=\hat{y}(x_j; \param)|x_{j} )      
    \end{align}
A perfectly calibrated model will have $\testacc(B_{r};\param) = \text{CONF}(B_{r};\param).$ Although Niculescu-Mizil et al.~\cite{niculescu2005predicting} noted that while DNNs are well-calibrated on binary classification tasks, it is impossible for a model to achieve perfect calibration in reality, making this another useful metric to assess the quality of models produced through different random seeds in conjunction with accuracy.

\section{Experiments}\label{sec: experiments}
Our proposed measure of model closeness/discrepancy offers new insight into questions around neural network training. In this section, we perform a series of illustrative experiments to demonstrate
the utility of this measure: %
\begin{itemize}
    \item In Section \ref{subsec: Ensemble Reference closeness}, we show that the eCDF of a deep ensemble model reliably approximates the reference as the size of the pool of candidate models in the ensemble increases. Establishing this relationship at the very onset allows us to qualitatively compare candidate models with the deep ensemble substitute of our reference function, while comparing candidate models to the reference function through the robust hypothesis test.
    \item In Section \ref{subsec: Minimum ensemble number}, we use our proposed measure of model closeness $\hat{\alpha}$ to show the minimum number of candidate models needed to form a reliable deep ensemble that has less variability in different performance metrics and approximates our reference function well. 
    \item In Section \ref{subsec: Discprepancy}, and \ref{subsec: Transfer Learning}, we connect the proposed discrepancy measure to other metrics used to measure network variability. We show how $\hat{\alpha}$ is more informative than validation accuracy alone through two case studies. The first uses a small CNN to perform a binary classification task on the CIFAR-10 dataset. The second uses a ViT variant, pre-trained on ImageNet, to perform the same task on the same dataset. The former case study allowed us to use neural networks with relatively few parameters, so we can test the limits of the expressivity of the proposed test and the latter to show how the same test applies to large-scale pre-trained models as well.
\end{itemize}
For experiments in Section \ref{subsec: Ensemble Reference Relation}, and \ref{subsec: Discprepancy}, we chose the following experimental setup. We used a small convolutional neural network with two convolutional layers (having $32$ and $16$ features, respectively, with a $3 \times 3$ kernel size) followed by one hidden layer of 64 units and a final layer of 2 units that output the raw logits of the network. Details of the network architecture and parameters are included in Section \ref{sec: CNN architecture} of Appendix. We train this network on a subset of the CIFAR-10 dataset~\cite{krizhevsky2009learning}, under the following settings:
    \begin{itemize}
        \item Of the 10 classes, we use 8 to create a binary classification problem by merging them into two super groups. Class 1 is comprised of airplane, automobile, ship, truck, and class -1 is comprised of bird, dog, frog and horse.
        \item The training size is $\Ntrain = 40000 $ and test size is $ \Ntest = 8000$.
        \item We fix all hyperparameters other than the random seeds by training all models for 50 epochs, with a fixed learning rate of 0.001, and fix the batch size to 32. 
    \end{itemize}
We chose a small example to allow us to train many models so that we can explore model training in different scenarios. The authors acknowledge the use of high-performance computing resources provided by the Office of Advanced Research Computing (OARC), at Rutgers, for running the experiments in this paper~\cite{Amarel}.
Code to compute the test statistic and $\hat{\alpha}$ metric of the two-sample version of the robust KS-test has been made available online~\cite{KSTestRepo}.

\subsection{Comparing the reference function and a deep ensemble}\label{subsec: Ensemble Reference Relation}

This section demonstrates how the eCDF of a deep ensemble model $\cdfavg$, defined in \eqref{eq:ensemble cdf}, closely approximates our reference function $\avgcdf$, defined in \eqref{eq:avg_cdf}. Establishing this relation will allow us to compare candidate models with the deep ensemble substitute of the reference function using different performance metrics like validation accuracy, churn w.r.t.~a deep ensemble, and the ECE, while comparing the eCDFs of candidate models to the reference function through the robust KS-test to analyze model variability. A lower ECE indicates a better-calibrated model, while lower churn w.r.t.~an ensemble indicates less disagreement between candidate models and an ensemble. We point the readers to Section \ref{sec: other metrics} for a formal definition of these metrics. 
\begin{figure}[tbh]
\begin{minipage}[b]{0.49\linewidth}
\centering
  \centerline{\includegraphics[width=4.2cm]{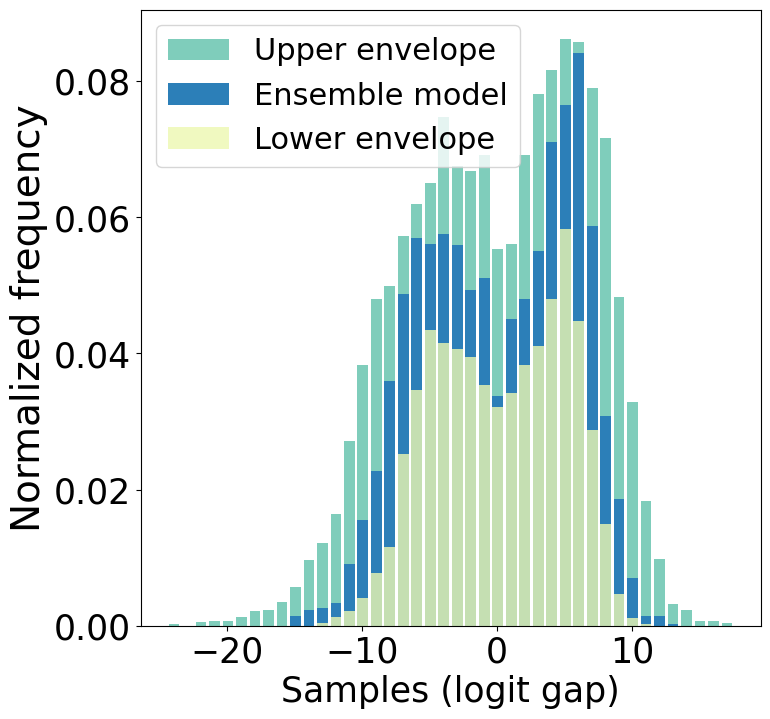}}
\end{minipage}
\begin{minipage}[b]{0.49\linewidth}
\centering
  \centerline{\includegraphics[width=4.5cm]{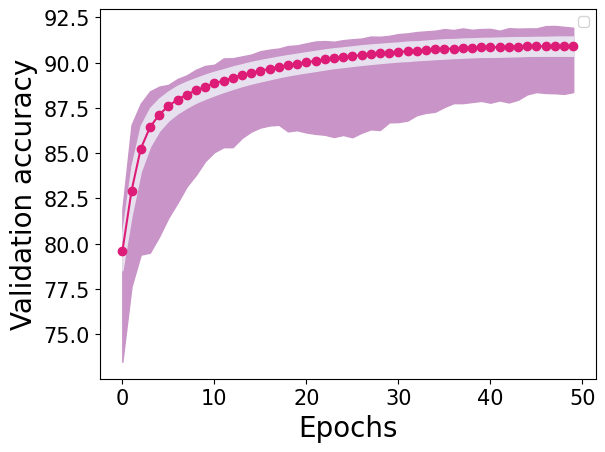}}
\end{minipage}
\caption{ (Left) Histogram of logit gaps from the ensemble model with the upper and lower envelopes representing the maximum and minimum probability attained in each bin among individual candidate models. (Right) A plot showing the evolution of validation accuracy of CNN models over different epochs. The solid red dots represent the mean validation accuracy over 800 seeds at each epoch, the light-colored area denotes one standard deviation, and the purple area represents the maximum and minimum values at that epoch.}
\label{fig: Hist logit gap ensemble}
\end{figure}
We train a total of $1600$ models by randomly fixing a seed for each model. The seeds control both initialization and batch order during SGD. We use the first $\Nmodels=800$ models to create a reference function $\avgcdf$ and set the rest $\Ncandmodels = 800$ models to create the eCDF of a deep ensemble $\cdfavg$. We choose different ensemble sizes, $\Nensemble \in [3, 5, 10, 20, 30, 70, 100, 150, 200]$, to observe the variability of ensembles in different metrics and to understand how large of an ensemble we need to approximate the reference function well. For each value of $\Nensemble$, we sample $\Nensemble$ models without replacement from the remaining $\Ncandmodels=800$ models and compute $\cdfavg$. We repeat this experiment $500$ times to get $500$ ensemble models using $\Nensemble$ components for each value of $\Nensemble$. Deep ensemble predictors have been widely used in the literature to reduce variability in DNN models~\cite{lakshminarayanan2017simple,lee2015m}. 
Candidate models will have varying degrees of certainty on individual test points. Taking the average of model ``confidences" across independent training runs makes them closer to their expected values, lowering variability. Although ensembling through averaging over softmax probabilities is common practice in the literature, averaging over logits has also been investigated to address the shortcomings of probability averaging~\cite{hinton2015distilling,wood2023unified}.

Figure \ref{fig: Hist logit gap ensemble} (Left) shows how the logit gap samples obtained from ensembling  $\Nensemble=\Ncandmodels$ candidate models compare with candidate models in the pool. The ensemble model produces fewer samples with small logit gaps (samples with higher uncertainty) and large logit gaps (overconfident samples).

Figure \ref{fig: Hist logit gap ensemble} (Right) shows the evolution of validation accuracy of candidate models over epochs. The solid red dots in the plot correspond to the mean accuracy over $\Ncandmodels$ seeds at each epoch, the light-colored region corresponds to one standard deviation, and the purple region corresponds to the minimum and maximum accuracy at that epoch. As observed in the plot, validation accuracy stops increasing from epoch 30 onwards, hence the decision to stop training at epoch 50, which is well past this optimization convergence. The same strategy was adopted by Picard~\cite[Section 4.1]{picard2021torch} who discusses it in more depth.

\subsubsection{Closeness of a deep ensemble to the reference function}\label{subsec: Ensemble Reference closeness}

The first question we want to answer is \textbf{how quickly does the eCDF of an ensemble model approximate the chosen reference?} In other words, as $\Nensemble \text{ approaches } \Ncandmodels$ how quickly does $\normifty{\avgcdf - \cdfavg}$ decrease?
To answer this, we compute the $L_{\infty}$-distance between $\avgcdf$ and $\cdfavg$ and compare it to the threshold set by the test in  \eqref{eq:dkw:cand}. To set the threshold $\delta_{a}$ in  \eqref{eq: Two samp threshold}, we fix $\epsilon_{a}=0.01$.

\begin{figure}[tbh]
\centering
  \centerline{\includegraphics[width=5cm]{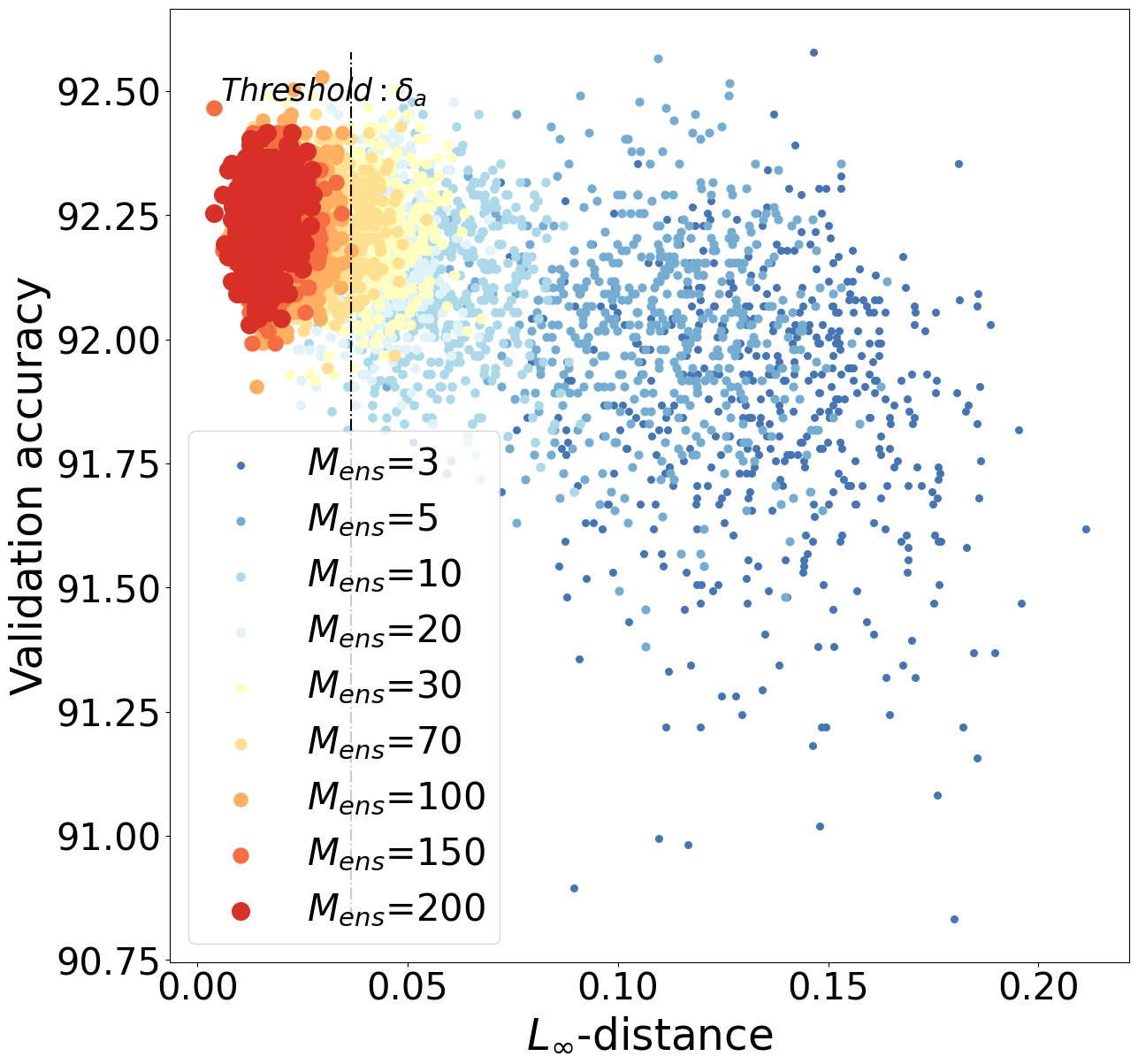}}
\caption{A plot showing $L_{\infty}$-distance of the eCDF of ensemble models $\cdfavg$ (formed with $\Nensemble$ models)  w.r.t.~the reference $\avgcdf$ (formed with $\Nmodels$ models) against validation accuracy of ensemble models. For each value of $\Nensemble$, we choose $\Nensemble$ candidate models to form one ensemble model and repeat this experiment 500 times through sampling with replacement to create 500 ensemble models. Thus, each dot with a fixed color represents one out of these 500 ensemble models and each color indicates the value of $\Nensemble$ or the number of candidate models in the pool. We chose $\epsilon=0.01$, to compute the threshold $\delta_{a}$ in  \eqref{eq: Two samp threshold}.}
\label{fig: Linfty_vstestacc}
\end{figure}

Figure \ref{fig: Linfty_vstestacc} demonstrates this convergence behavior. The eCDF of all ensemble models $\cdfavg$, produced with $\Nensemble=100, 150, \text{and } 200$ candidate models have $L_{\infty}$-distances from $\avgcdf$ less than the threshold set for our KS-test. This empirical evidence indicates that a deep ensemble eCDF more closely resembles the reference function $\avgcdf$ as $\Nensemble \text{ approaches } \Ncandmodels$. Therefore, we expect any candidate model close to our reference function $\avgcdf$, formed with $\Nmodels$ candidate models, to be also close to a deep ensemble, formed with $\Nensemble$ candidate models, in terms of different performance metrics, provided $\Nensemble \text{ approaches } \Ncandmodels=\Nmodels$.

\begin{figure*}[hbt]
 \begin{minipage}[b]{0.28\linewidth}
\centering
  \centerline{\includegraphics[width=5.3cm]{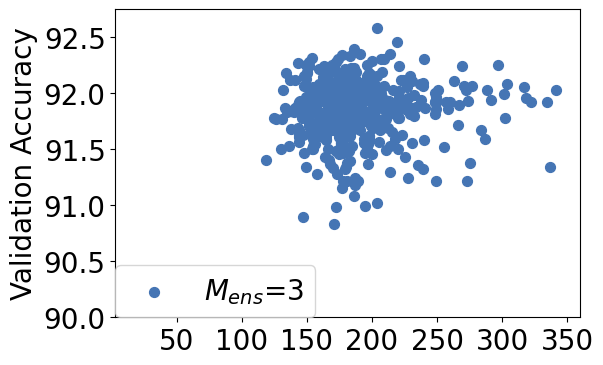}}
\end{minipage}
\begin{minipage}[b]{0.23\linewidth}
\centering
  \centerline{\includegraphics[width=4.4cm]{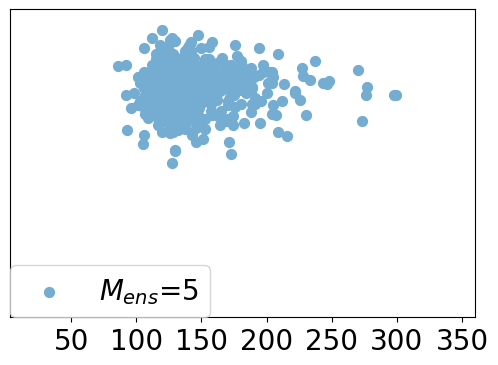}}
\end{minipage}
\centering
 \begin{minipage}[b]{0.23\linewidth}
\centering
  \centerline{\includegraphics[width=4.4cm]{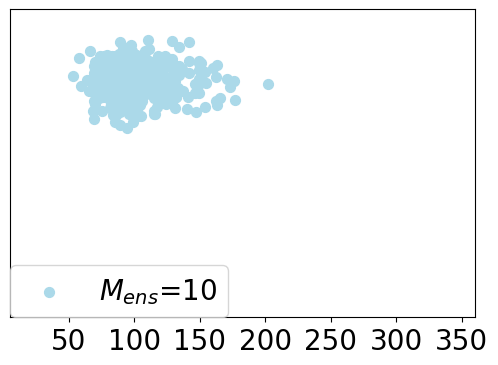}}
\end{minipage}
\begin{minipage}[b]{0.23\linewidth}
\centering
  \centerline{\includegraphics[width=4.4cm]{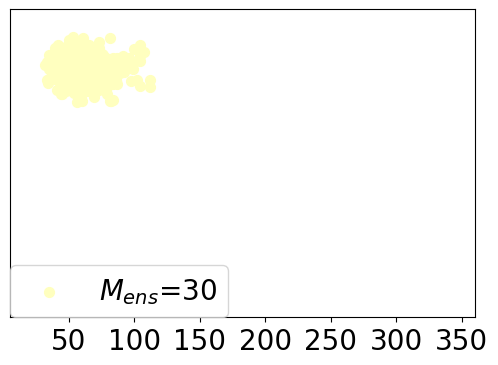}}
\end{minipage}
 \begin{minipage}[b]{0.28\linewidth}
\centering
  \centerline{\includegraphics[width=5.3cm]{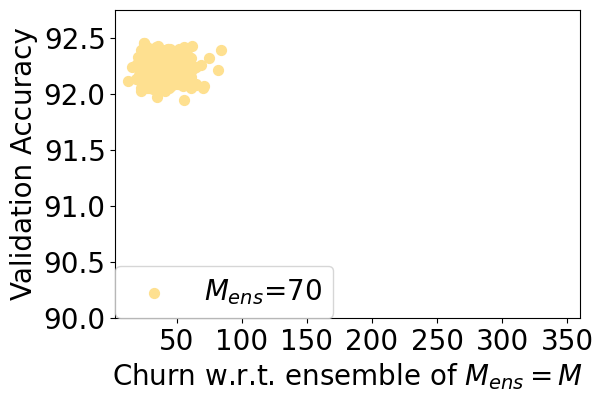}}
\end{minipage}
\begin{minipage}[b]{0.23\linewidth}
\centering
  \centerline{\includegraphics[width=4.4cm]{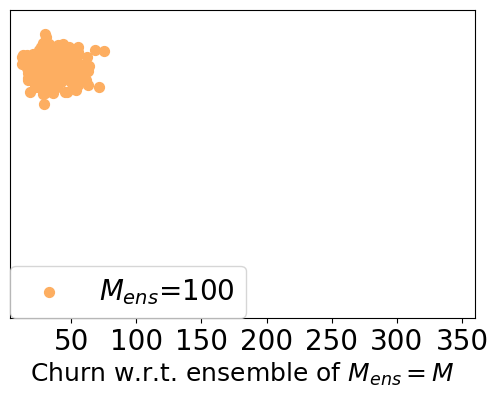}}
\end{minipage}
\centering
 \begin{minipage}[b]{0.23\linewidth}
\centering
  \centerline{\includegraphics[width=4.4cm]{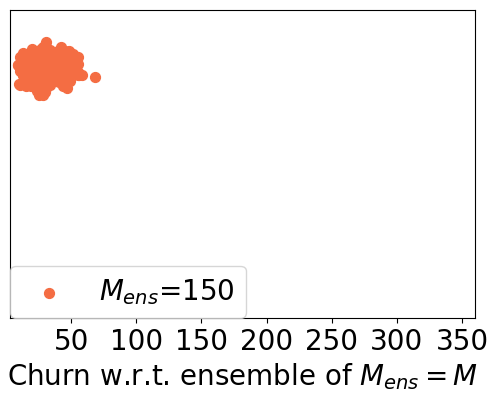}}
\end{minipage}
\begin{minipage}[b]{0.23\linewidth}
\centering
  \centerline{\includegraphics[width=4.4cm]{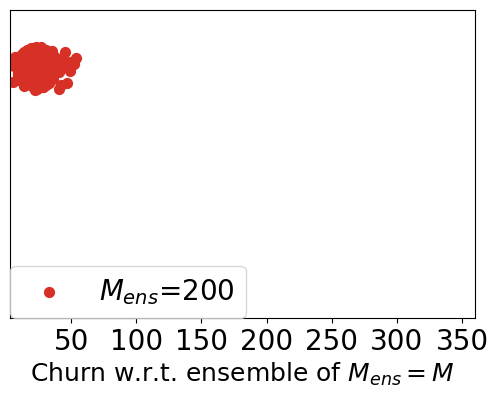}}
\end{minipage}

\caption{2-D scatter plot to visualize the relationship among ensemble models in terms of validation accuracy, and churn w.r.t.~an ensemble of size $\Nensemble=\Nmodels$. The ensemble models are the same as Figure \ref{fig: Linfty_vstestacc}.}\label{fig: Testacc churn ensembles}
\end{figure*}

\begin{figure*}[hbt]
 \begin{minipage}[b]{0.255\linewidth}
\centering
  \centerline{\includegraphics[width=5cm]{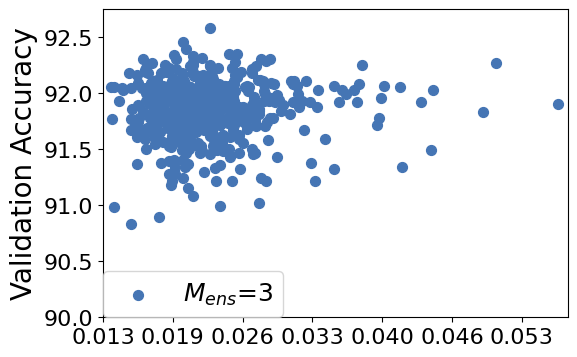}}
\end{minipage}
\begin{minipage}[b]{0.248\linewidth}
\centering
  \centerline{\includegraphics[width=4.45cm]{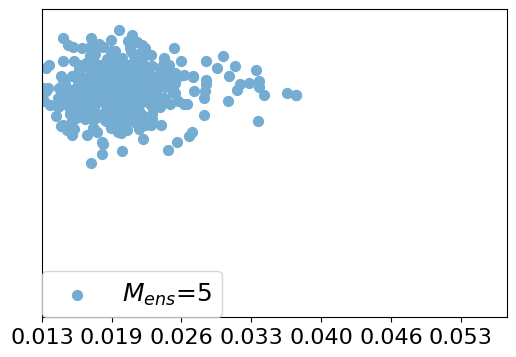}}
\end{minipage}
\centering
 \begin{minipage}[b]{0.23\linewidth}
\centering
  \centerline{\includegraphics[width=4.45cm]{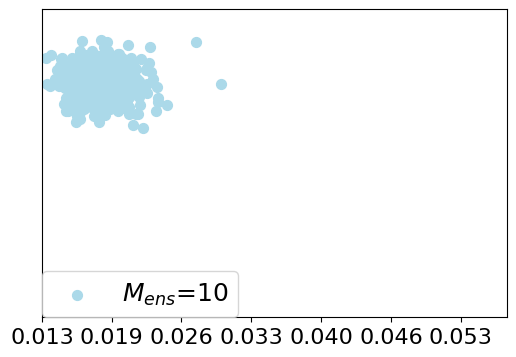}}
\end{minipage}
\begin{minipage}[b]{0.24\linewidth}
\centering
  \centerline{\includegraphics[width=4.45cm]{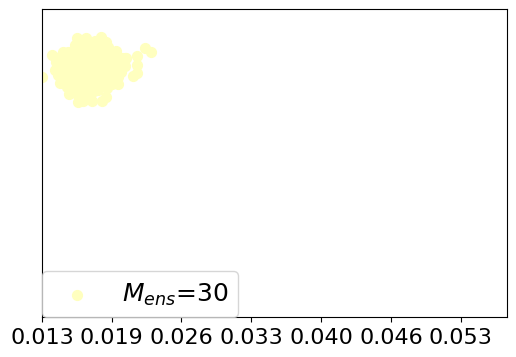}}
\end{minipage}
 \begin{minipage}[b]{0.255\linewidth}
\centering
  \centerline{\includegraphics[width=5cm]{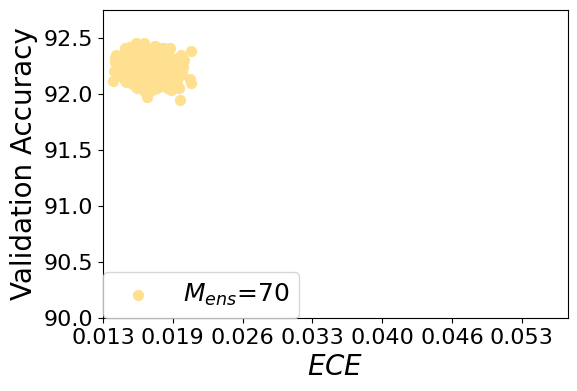}}
\end{minipage}
\begin{minipage}[b]{0.248\linewidth}
\centering
  \centerline{\includegraphics[width=4.45cm]{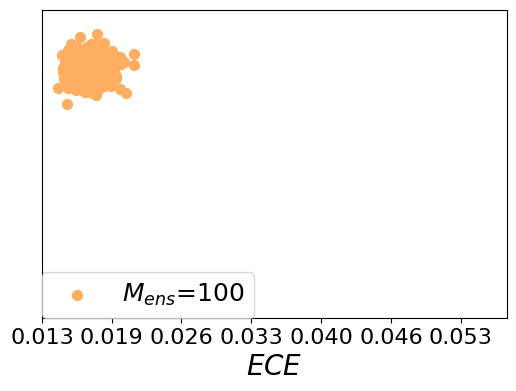}}
\end{minipage}
\centering
 \begin{minipage}[b]{0.23\linewidth}
\centering
  \centerline{\includegraphics[width=4.45cm]{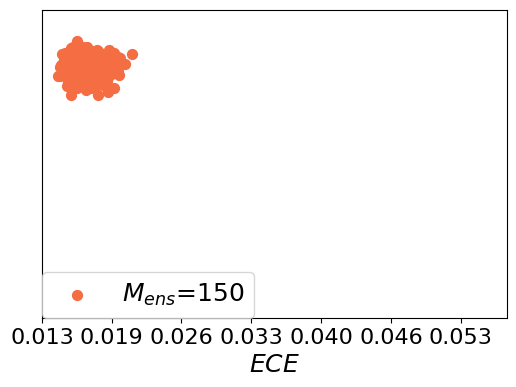}}
\end{minipage}
\begin{minipage}[b]{0.24\linewidth}
\centering
  \centerline{\includegraphics[width=4.45cm]{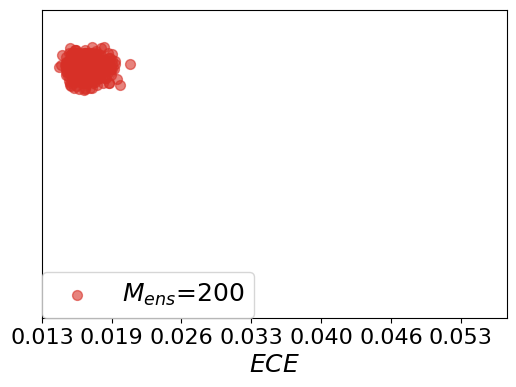}}
\end{minipage}

\caption{2-D scatter plot to visualize the relationship among ensemble models in terms of validation accuracy, and ECE. The ensemble models are the same as Figure \ref{fig: Linfty_vstestacc}.}\label{fig: Testacc ece ensembles}
\end{figure*}

Since the benefits of ensembling are observed even for pool sizes as small as $\Nensemble =5$~\cite{lakshminarayanan2017simple}, it might be tempting to focus on a few random seeds to generate a deep ensemble model due to computational constraints. This brings us to the next question: \textbf {Does averaging over a small number of candidate models to create an ensemble model result in a reliable representative of the training procedure?} We try to understand this qualitatively. We investigate three metrics of ensemble models as $\Nensemble$ approaches $\Ncandmodels$: validation accuracy, churn w.r.t.~a deep ensemble of pool size $\Nensemble=\Nmodels$, and ECE.
In Figure \ref{fig: Testacc churn ensembles}, and  Figure \ref{fig: Testacc ece ensembles}, we use 2-D scatter plots to visualize the relation among the ensemble models w.r.t.~these three metrics. We expect the best ensembles to have high validation accuracy, low churn, and low ECE. 

As observed in Figure \ref{fig: Testacc churn ensembles} and \ref{fig: Testacc ece ensembles}, as we increase $\Nensemble$, we notice the ensemble models converge to high validation accuracy, low churn, and low ECE. These ``good" ensembles are also the same models whose eCDFs have achieved a smaller $L_{\infty}$-distance from the reference function $\avgcdf$ in Figure \ref{fig: Linfty_vstestacc}. Some ensembles created from a small number of models may outperform those created from a large number of models in one particular metric, but it is always at the cost of performance w.r.t.~another metric. For instance, one of the highest validation accuracies achieved in Figure \ref{fig: Testacc churn ensembles} is an ensemble model belonging to $\Nensemble=3$. However, this model also has a higher churn w.r.t.~an ensemble of size $\Nensemble=\Nmodels$ compared to ensembles models with large $\Nensemble$ ($\Nensemble = 100, 150, 200)$. This is also reflected by the high $L_{\infty}$-distance of this model from the reference $\avgcdf$ as shown in Figure \ref{fig: Linfty_vstestacc}. We conclude that contrary to standard practice, exploring more than 5 seeds is important to create a reliable ensemble that has less variability in different performance metrics. Thus, \textbf{ensembling over smaller pool sizes results in a deep ensemble that has significant variability over different performance metrics, and increasing the size $\Nensemble$ of ensemble models reduces this variability.}

\subsubsection{Selecting the number of models to be ensembled via the robust KS-test}\label{subsec: Minimum ensemble number}
\begin{figure}[thb]
\centering
  \centerline{\includegraphics[width=5cm]{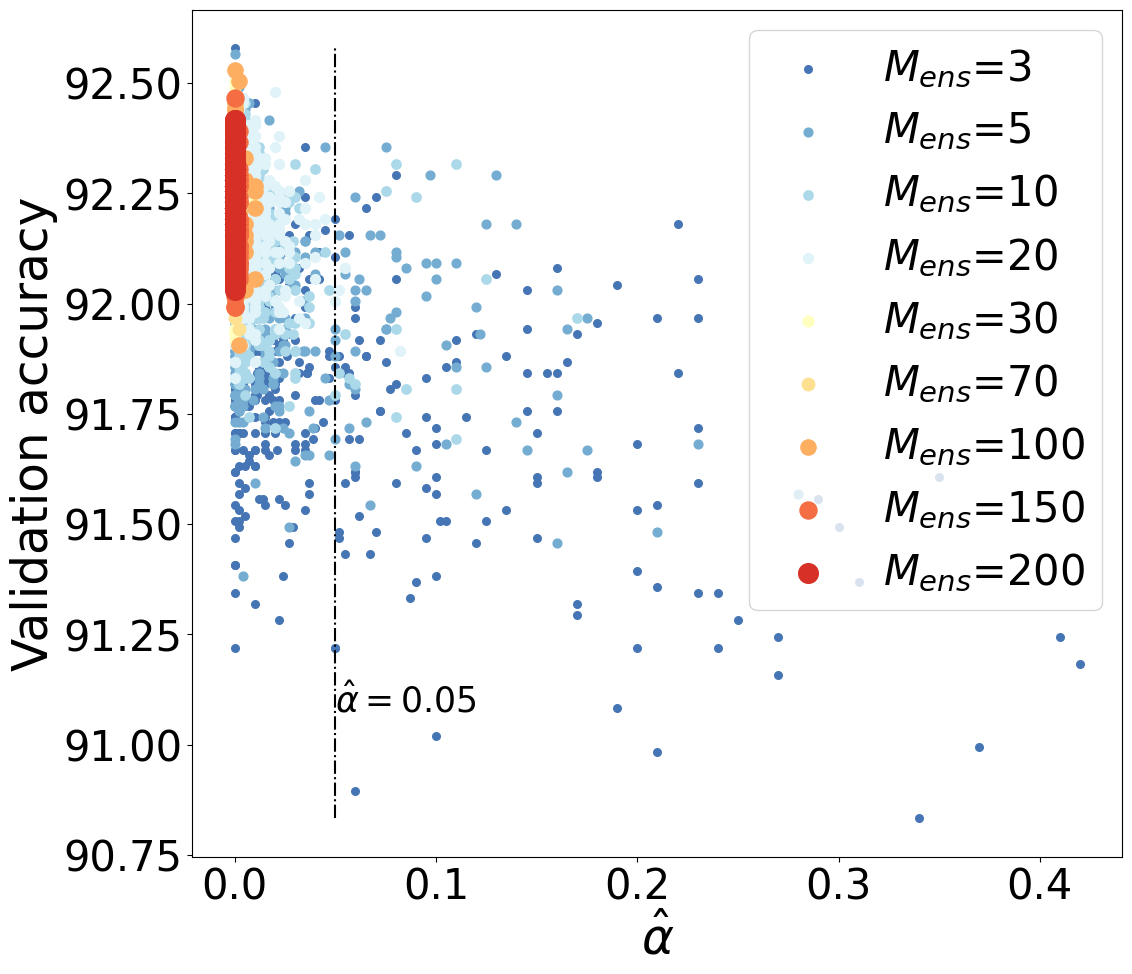}}
\caption{A plot showing $\hat{\alpha}$ of the eCDF of ensemble models $ \cdfavg$ (formed with $\Nensemble$ models)  w.r.t.~the reference $\avgcdf$ (formed with $\Nmodels$ models) against validation accuracy of ensemble models. We randomly select $\Nensemble$ candidate models for each value of $\Nensemble$ to form an ensemble. We repeat this procedure 500 times, sampling with replacement, to create 500 ensemble models. Thus, each dot with a fixed color represents one out of these 500 ensemble models and each color indicates the value of $\Nensemble$ or the number of candidate models in the pool. The ensemble models are the same as Figure \ref{fig: Linfty_vstestacc}.}
\label{fig: alphahat_vstestacc}
\end{figure}

Section \ref{subsec: Ensemble Reference closeness} showed that it is possible to create a good ensemble with enough models that have less variability over different metrics and whose eCDF approximates the reference function well. The more models we ensemble, the better and more consistent the ensembles become. This brings us to our final question: \textbf{How many candidate models do we need to create a deep ensemble model that is a reliable representative of the training procedure i.e. has less variability over different 
performance metrics and approximates the reference function well?} Based on the threshold set by our test on the $L_{\infty}$-distance and as demonstrated in Figure \ref{fig: Linfty_vstestacc}, a classical KS-test will indicate that we need to explore 100 random seeds to create a reliable ensemble, which can then be used as a base model for qualitative comparisons with candidate models.  Since the classical KS-test is too sensitive, setting a threshold based on it will require too many models to create the ensemble. We can use the proposed robust KS-test to also measure the similarity between the ensemble function $\cdfavg$ and the reference function $\avgcdf$ but with a little wiggle room. In practice, we also may not want to rely on validation accuracy or churn to decide on the number of models we need. So, the proposed test offers a way to estimate this number using only the logits. To that end, we compute $\hat{\alpha}$ through a robust KS-test by measuring the closeness/discrepancy of the eCDFs of ensemble models $\cdfavg$, of different sizes, with the reference function $\avgcdf$. Similar to Figure \ref{fig: Linfty_vstestacc}, in Figure \ref{fig: alphahat_vstestacc}, as $\Nensemble$ approaches $\Ncandmodels$, $\hat{\alpha}$ approaches 0. However, if we set a threshold on the $\alpha$-trimming level instead of the $L_{\infty}$-distance, we get a better lower bound on the number of models we need to use for a reliable ensemble. If we consider all ensembles that require $\hat{\alpha}\le 0.05$, we see that ensembles created from $\Nensemble=30$ models or more are all included in this set as shown in column 2 of Table \ref{Table: Ensemble stats}, while more than $20 \%$ of ensembles created from $\Nensemble=3$ models have required a higher level of trimming. From Table \ref{Table: Ensemble stats}, we notice that ensembles achieve better values in all metrics as the size of the ensemble $\Nensemble$ increases and the variability of these three metrics reduces noticeably for ensembles created from $\Nensemble=30$ models or more. \textbf{Thus, our proposed metric indicates that ensembling over at least $\Nensemble=30$ candidate models may be required for an ensemble to be a reliable representative of the training variability.
 }

\begin{table}[htb]
\caption{Ensemble statistics for different values of $\Nensemble$}
\label{Table: Ensemble stats}
\centering
\resizebox{\columnwidth}{!}{\begin{tabular}{|c|c|c|c|c|}
 \hline
 $\mathbf{\Nensemble}$ & \textbf{\% of models} & \textbf{Accuracy} & \textbf{Churn w.r.t.~ensemble} & \textbf{ECE } \\
   & \textbf{with} $\mathbf{\hat{\alpha} \le 0.05}$ & \textbf{mean $\pm$ std} & \textbf{mean $\pm$ std} &  \textbf{mean $\pm$ std} \\
 \hline
   3 & 77.20 & $91.85 \pm 0.25$ & $190.79 \pm 33.90$ & $0.0226 \pm 0.0048$   \\
 \hline
   5 & 89.20 & $92.03 \pm 0.18$  & $145.70 \pm 28.61$  &  $0.0200 \pm 0.0037$\\
 \hline
  10 & 96.80  & $92.11 \pm 0.14$ & $103.57 \pm 22.50$ & $0.0185 \pm 0.0023$ \\
  \hline
   20  & 99.40  & $92.17 \pm 0.11$ & $76.972 \pm 17.23$ & $0.0175 \pm 0.0016$ \\
  \hline   
   30 &  100.00  & $92.21 \pm 0.10$ & $62.84 \pm 12.45$ &  $0.0171 \pm 0.0014$ \\
 \hline
  70 &  100.00  &  $92.22 \pm 0.08$ & $43.45 \pm 9.64$ & $0.0168 \pm 0.0012$\\
 \hline
  100 & 100.00  & $92.23 \pm 0.08$ & $38.59 \pm  9.08$ & $0.0169 \pm 0.0012$\\
 \hline
 150 & 100.00   &  $92.23 \pm 0.07$ & $32.75 \pm 7.42$  & $0.0169 \pm 0.0011$ \\
 \hline
 200 & 100.00  &  $92.24 \pm 0.07$ & $29.3 \pm 6.91$ & $0.0169 \pm 0.0011$ \\
 \hline
\end{tabular}}
\end{table}

\subsection{Evaluating the proposed metric of model closeness/ discrepancy}\label{subsec: Discprepancy}
We next move on to understand how our proposed measure of model closeness/discrepancy relates to metrics commonly used to understand model variability. We try to understand this for candidate models generated under different sources of randomness ( like initialization and batch order). 
In particular, we considered three scenarios: $\randinit$ with only random initialization and fixed batches, $\randbatch$ with only random batch selection in SGD and fixed initialization, and $\randall$ combining both sources of randomness. We trained 200 models in each of $\randinit$, $\randbatch$, and  $\randall$ with the same hyperparameter settings as detailed at the beginning of Section \ref{sec: experiments}. For each source of randomness, we use $\Nmodels = 100$ models to create the reference function $\avgcdf$, as defined in  $\eqref{eq:avg_cdf}$ and from the rest $\Ncandmodels = 100$ models, we choose our candidate models. For each candidate model, we use its eCDF $\multcandcdf$, $l \in [\Ncandmodels]$,  as defined in  $\eqref{eq:Individual ecdf}$, to compute $\hat{\alpha}$ through a robust KS-test against the reference $\avgcdf$.

\subsubsection{Comparing our proposed discrepancy measure with accuracy}
 \begin{figure}[htb]
\centering
  \centerline{\includegraphics[width=8.5cm]{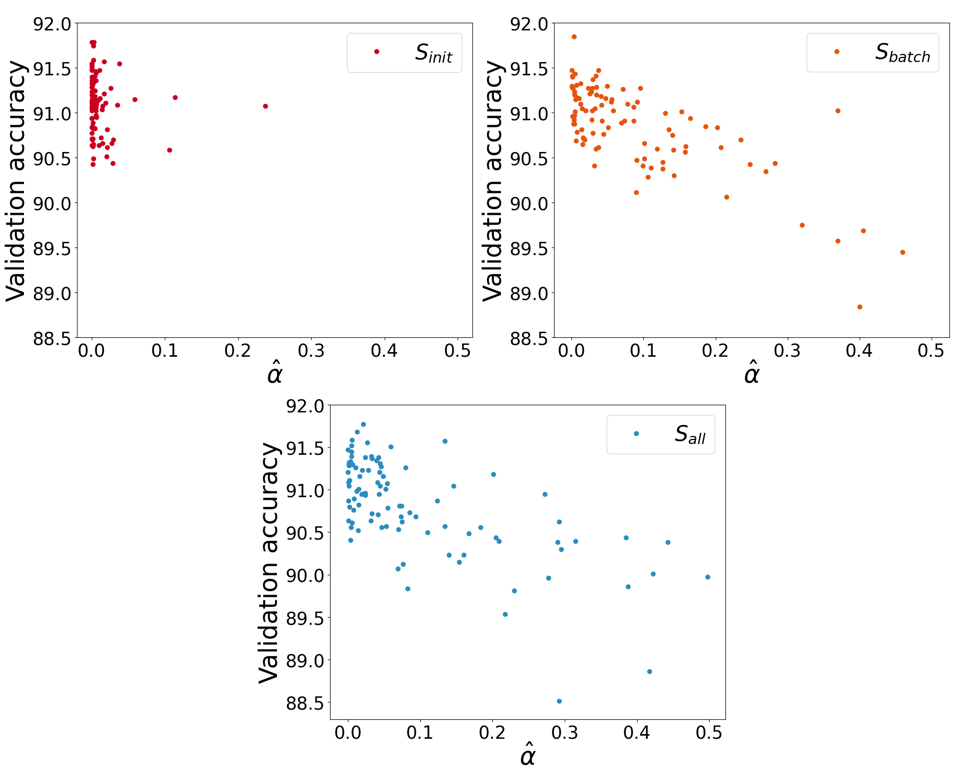}}
 \caption{2-D Scatter plot to visualize how $\hat{\alpha}$, computed using the robust KS-test against the reference $\avgcdf$, relates to validation accuracy for 100 candidate models (each dot in the plots) in $\randinit$, $\randbatch$ and $\randall$.} 
\label{fig: Testacc_vs_alpha}
\end{figure}
Our first question in this section is: \textbf{How does our proposed metric $\hat{\alpha}$ relate to accuracy?} In Figure \ref{fig: Testacc_vs_alpha}, we plot the relationship between the validation accuracy of candidate models and $\hat{\alpha}$ for candidate models in $\randinit$, $\randbatch$, and $\randall$. For each plot, we notice significant variability over random seeds in both $\hat{\alpha}$ and validation accuracy for the same hyperparameter setting. This is evidence that a fixed hyperparameter setting that works well for one random seed can perform poorly for another and thus produce models with very different logit gap functions. We notice more variability in both validation accuracy and $\hat{\alpha}$ among models in $\randbatch$, than models in $\randinit$. Details of accuracy statistics for each source of randomness are listed in Table \ref{tab:Validation accuracy stats random sources}. 
\begin{table}[htb]
\caption{Validation accuracy from different sources of randomness}
\label{tab:Validation accuracy stats random sources}
\centering   
\resizebox{\columnwidth}{!}{\begin{tabular}{|c|c|c|c|c|}
\hline
\textbf{Source of} & \textbf{Accuracy} & \textbf{Accuracy} & \textbf{Ensemble} &\textbf{Accuracy range of} \\
 \textbf{randomness} & \textbf{mean $\pm$ std} & \textbf{range} & \textbf{ accuracy}& \textbf{models with} $\mathbf{\hat{\alpha} \le 0.05}$\\
\hline
 $\randinit$ & $91.101 \pm 0.302$ & [90.422, 91.781] & 91.731 & [90.422, 91.781] \\ 
 \hline
$\randbatch$ & $90.819 \pm 0.554$ & [87.852, 91.843]  & 91.731 & [90.409, 91.843]  \\
 \hline
 $\randall$ & $90.810 \pm 0.560$ & [88.513,  91.769]  & 92.361 & [90.409, 91.769] \\
 \hline
\end{tabular}}
\end{table}
The mean accuracy of models in $\randinit$ is slightly higher than models in $\randall$, and $\randbatch$, while all three categories have achieved similar maximum accuracy as observed in the accuracy range column. The higher variability among models in $\randbatch$ comes from producing models with much poorer validation accuracy than $\randinit$, a trend also observed in $\randall$, as reported in the minimum accuracy of the accuracy range for each category. This indicates that in this case study, random batch shuffling has a higher effect on the overall variability among models in $\randall$, than random initialization. We also notice that although there is more variability among models in $\randbatch$, than $\randinit$, the ensemble model has performed the same for both categories. However, the combined variability of $\randinit$, and $\randbatch$, reflected among models in $\randall$ has led to a better ensemble performance in $\randall$.

We also look at models admitted by small values of $\hat{\alpha}$ in each category (e.g. $\hat{\alpha} \le 0.05$), and observe the corresponding range of validation accuracy, reported in the fourth column of Table \ref{tab:Validation accuracy stats random sources}. For $\randinit$, the range of accuracy achieved by models admitted by $\hat{\alpha} \le 0.05$ covers the full range of accuracy of all models in $\randinit$. This range is also close to the range of models admitted by $\hat{\alpha} \le 0.05$ in $\randall$, and $\randbatch$. The maximum validation accuracy for each source is also admitted by small values of $\hat{\alpha}$.
This is because candidate models aggregate to form ensembles that result in a boost over average performance in the group. Since models with lower $\hat{\alpha}$ are close to the reference function in terms of their eCDF, and hence also to the ensemble, they will also have accuracy similar to the ensemble. For each source of randomness, if we consider the range of validation accuracy for models admitted by small values of $\hat{\alpha}$ to be representative of the training variability (as a result of being close to the reference function), then any model that has performed worse than this range will end up with a high value of $\hat{\alpha}$. This can be observed for $\randbatch$, and $\randall$ since these two categories have resulted in more models with performance worse than the discussed range. However, the opposite is not true, i.e. models with accuracy within the discussed range will not always have a small $\hat{\alpha}$, and hence will not be good representatives. This shows that validation accuracy alone is not the right metric to assess model quality. A model can end up within the discussed range of validation accuracy but not be a good representative of the training variability over random seeds. We conclude that \textbf{smaller values of $\hat{\alpha}$ does not admit poor validation accuracy, but similar validation accuracy does not imply similar $\hat{\alpha}$}.
\subsubsection{Comparing our proposed discrepancy measure with pairwise churn}
\begin{figure*}[hbt]
 \begin{minipage}[b]{0.265\linewidth}
\centering
  \centerline{\includegraphics[width=5.05 cm]{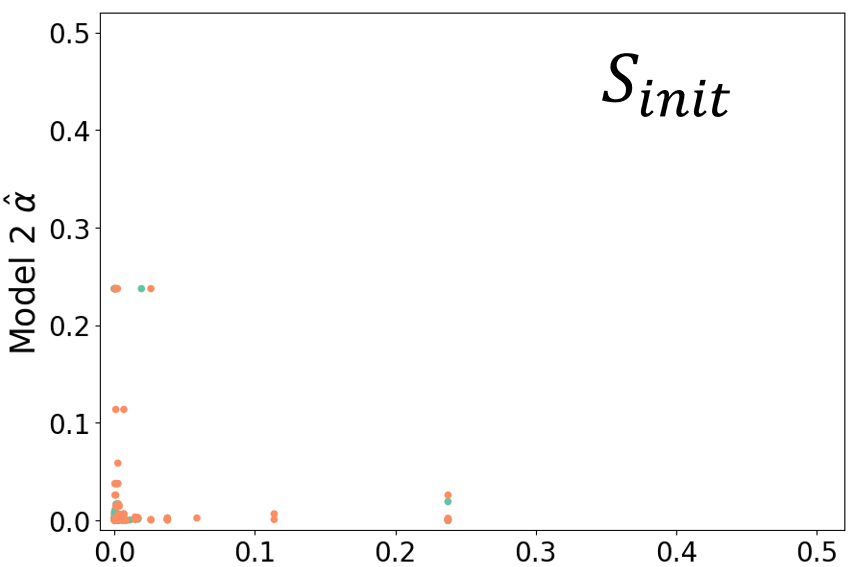}}
\end{minipage}
\begin{minipage}[b]{0.245\linewidth}
\centering
  \centerline{\includegraphics[width=4.45cm]{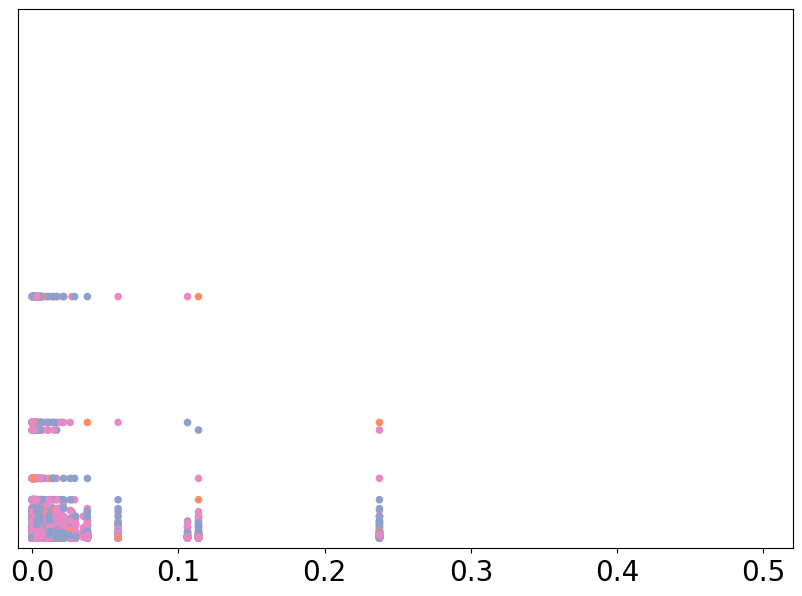}}
\end{minipage}
\centering
 \begin{minipage}[b]{0.24\linewidth}
\centering
  \centerline{\includegraphics[width=4.45cm]{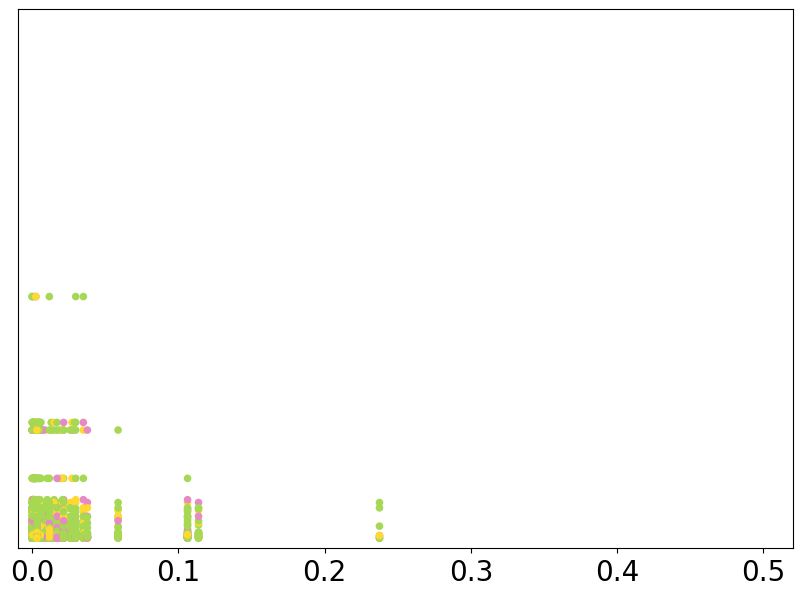}}
\end{minipage}
 \begin{minipage}[b]{0.23\linewidth}
\centering
  \centerline{\includegraphics[width=4.2cm]{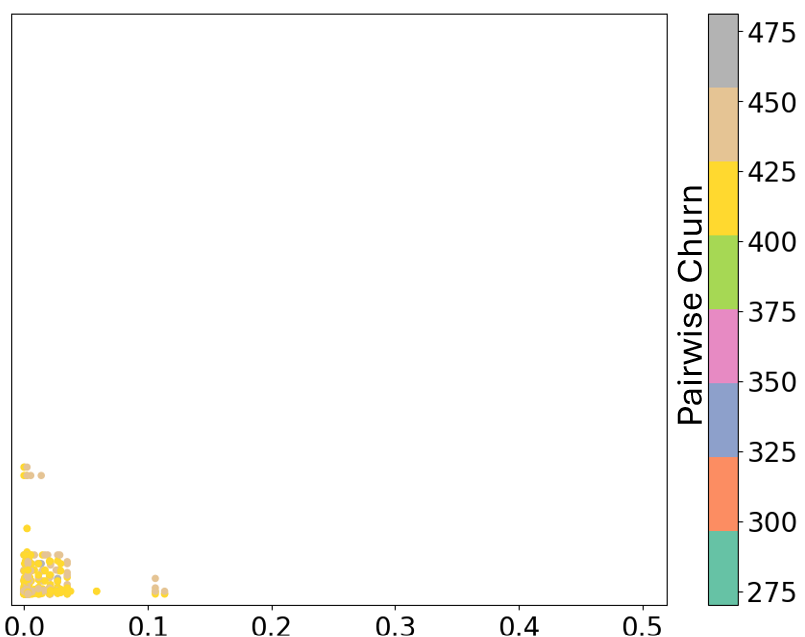}}
\end{minipage}
\begin{minipage}[b]{0.265\linewidth}
\centering
  \centerline{\includegraphics[width=5.05cm]{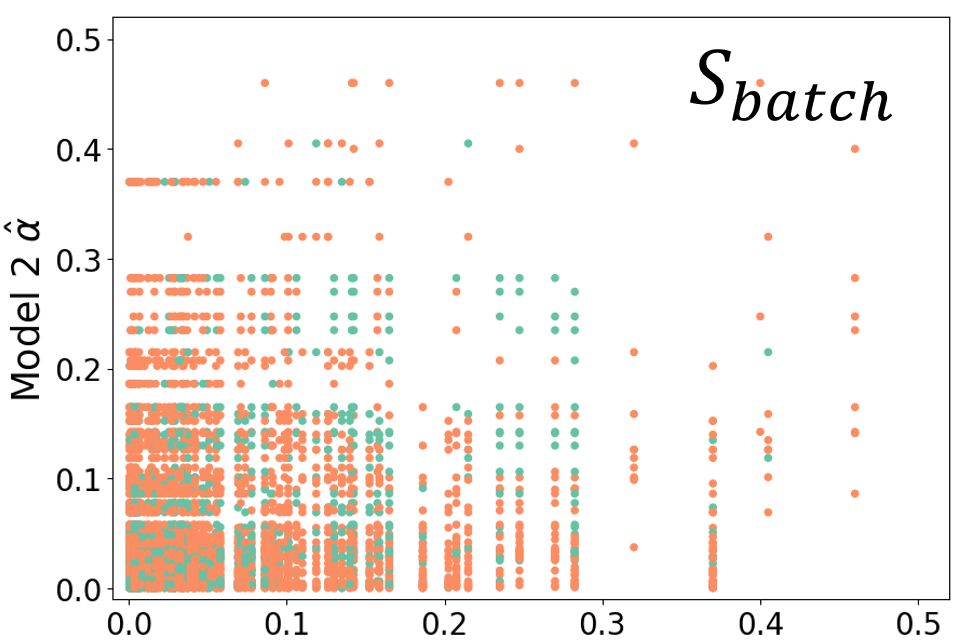}}
\end{minipage}
\centering
 \begin{minipage}[b]{0.245\linewidth}
\centering
  \centerline{\includegraphics[width=4.45cm]{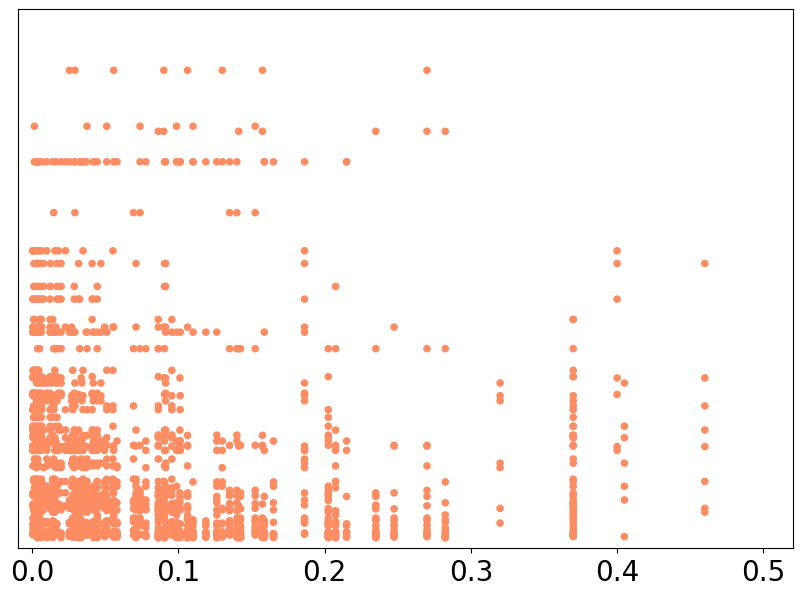}}
\end{minipage}
 \begin{minipage}[b]{0.24\linewidth}
\centering
  \centerline{\includegraphics[width=4.45 cm]{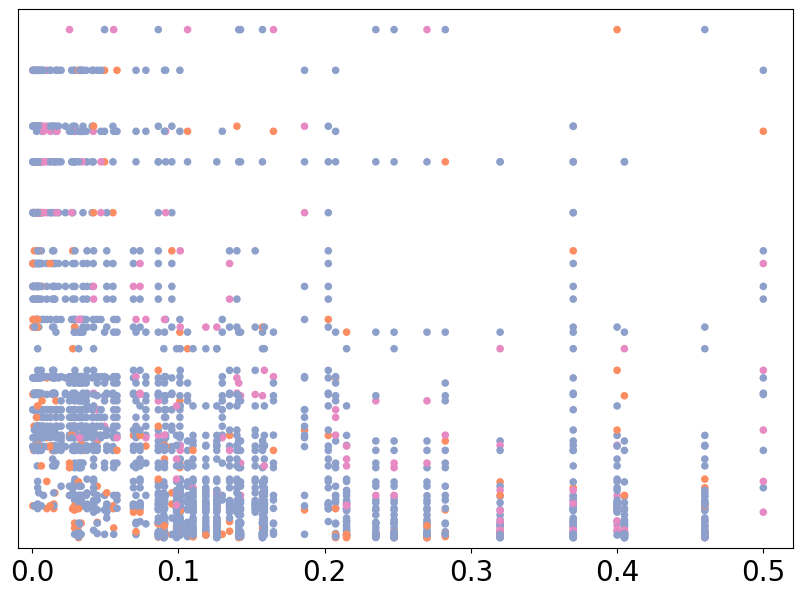}}
\end{minipage}
\begin{minipage}[b]{0.23\linewidth}
\centering
  \centerline{\includegraphics[width=4.2cm]{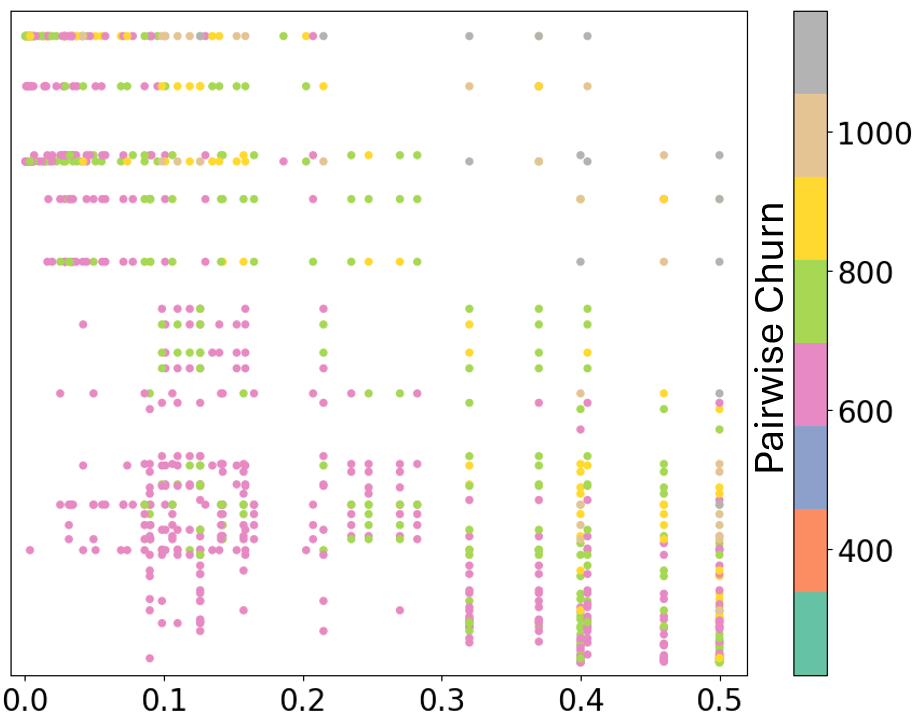}}
\end{minipage}
\centering
 \begin{minipage}[b]{0.265\linewidth}
\centering
  \centerline{\includegraphics[width=5.0cm]{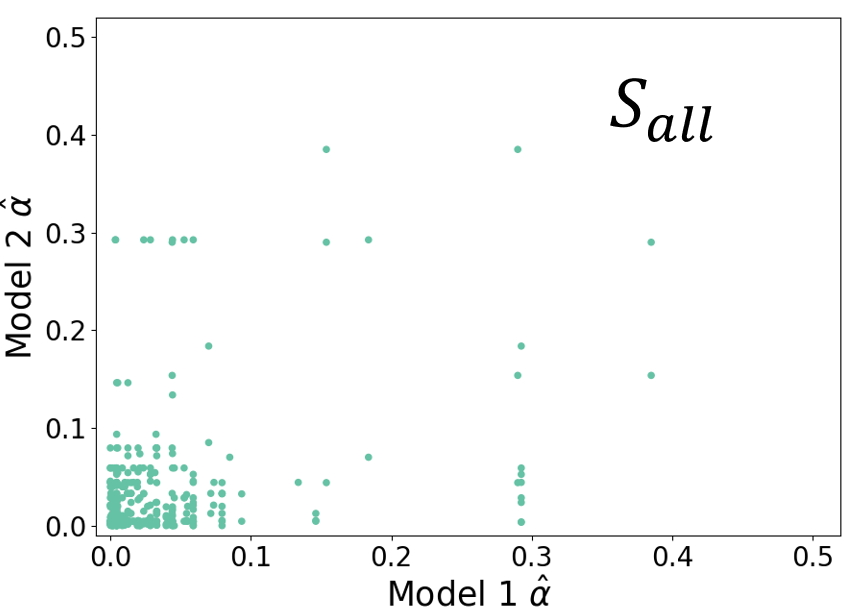}}

\end{minipage}
 \begin{minipage}[b]{0.245\linewidth}
\centering
  \centerline{\includegraphics[width=4.45 cm]{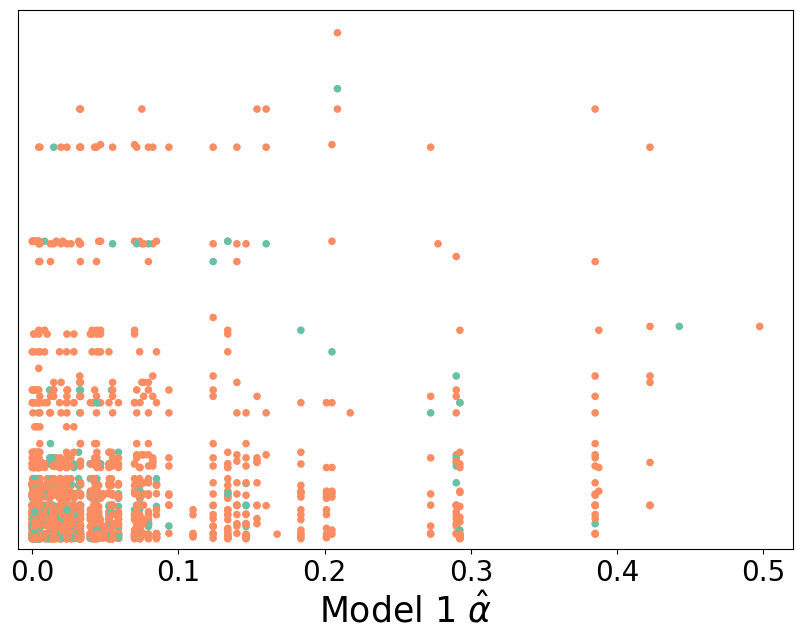}}
\end{minipage}
 \begin{minipage}[b]{0.24\linewidth}
\centering
  \centerline{\includegraphics[width=4.45 cm]{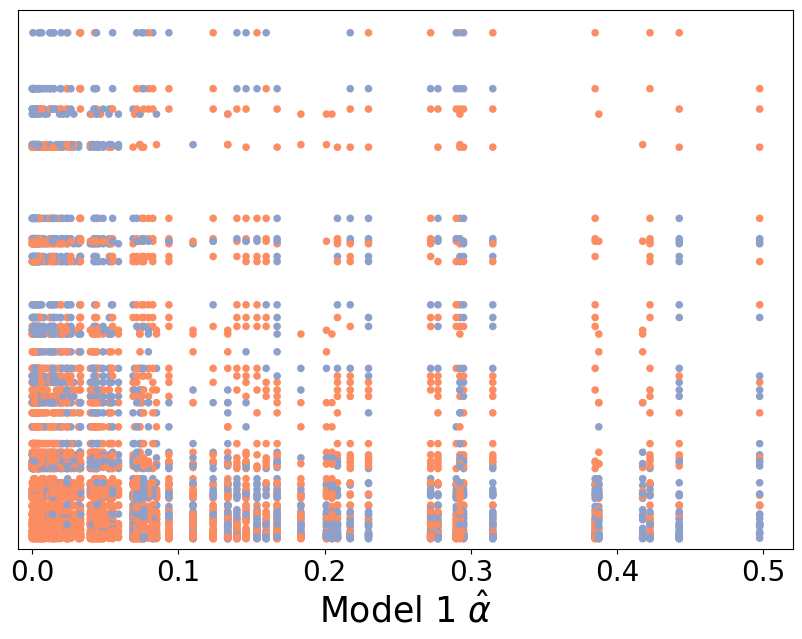}}
\end{minipage}
 \begin{minipage}[b]{0.23\linewidth}
\centering
  \centerline{\includegraphics[width=4.2 cm]{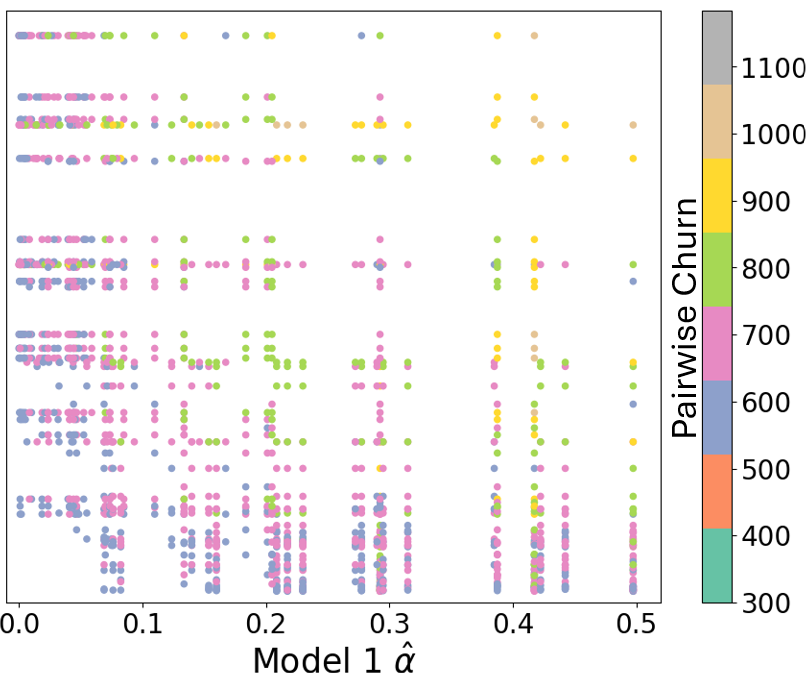}}
\end{minipage}
\caption{2-D Scatter plot to visualize how pairwise values of $\hat{\alpha}$, computed using the robust KS-test against the reference $\avgcdf$, relates to pairwise churn for each pair of $(100\times100)$ candidate models (each dot in the plots). 
For each source of randomness (plotted along the rows), the total pairwise churn is divided into 4 ranges (plotted along the columns) indicated by the colorbar. The banded structure of the plot is because our hypothesis test checks for only discrete values of $\alpha$.
}
\label{fig:PairwiseChurn_vs_alpha}
\end{figure*}
\begin{table}[htb]
\caption{Pairwise churn for different types of randomness}
\label{tab: pairwise churn stats}
\centering   
\resizebox{\columnwidth}{!}{\begin{tabular}{|c|c|c|}
\hline
\textbf{Source of} & \textbf{Range of} & \textbf{ Range of pairwise churn of pairs of}\\
\textbf{randomness} & \textbf{pairwise churn} & \textbf{models with $(\mathbf{\hat{\alpha_{i}} \le 0.05},\mathbf{\hat{\alpha_{j}} \le 0.05}), i \neq j$}\\
\hline
 $\randinit$ & [270, 481 ] & [270, 481] \\ 
 \hline
$\randbatch$  & [218, 1174 ] & [ 218, 538]  \\
 \hline
 $\randall$ & [327, 1184 ] &  [327, 591] \\
 \hline
\end{tabular}}
\end{table}
\textbf{How does our proposed metric $\hat{\alpha}$ relate to pairwise churn?} In Figure \ref{fig:PairwiseChurn_vs_alpha}, we plot the relationship between the pairwise disagreement between candidate models or churn, and $\hat{\alpha}$, for candidate models in $\randinit$, $\randbatch$, and $\randall$. We distinguish this from churn w.r.t.~a deep ensemble by labeling it as pairwise churn. Similar to the trend observed for validation accuracy, there is more variability in pairwise churn among pairs of models in $\randbatch$, than $\randinit$, as observed in the pairwise churn range for each random source reported in column 2 of Table \ref{tab: pairwise churn stats}. Again, most of the variability in pairwise churn in $\randall$ is due to $\randbatch$. 
In Figure \ref{fig:PairwiseChurn_vs_alpha}, for $\randbatch$, and $\randall$, we observe that as the range of pairwise churn increases, the scatter plot becomes less dense in the region where both the models have a low $\hat{\alpha}$. This indicates that pairs of models with low values of $\hat{\alpha}$ for both models achieve pairwise churn in the smaller range among its group. This is less obvious for models in $\randinit$, because this group's total range of pairwise churn is low. We see this in column 3 of Table \ref{tab: pairwise churn stats}, where we focus on pairs of models in each category that have achieved pairwise $\hat{\alpha}$ less than  $0.05$ and report the range of pairwise churn for those models. Since the range of pairwise churn of pairs of models in $\randinit$ is low, small values of pairwise $\hat{\alpha}$ have admitted this full range. However, for $\randbatch$, and $\randall$, since the total range of pairwise churn is much larger than that of $\randinit$, small values of $\hat{\alpha}$ have admitted only those pairs of models that have achieved a pairwise churn in the smaller range of the total range. Any pairs of models with pairwise churn higher than this range will have at least one model in the pair with a high $\hat{\alpha}$. A possible explanation for a pair of models with low pairwise $\hat{\alpha}$ also having a low pairwise churn is that models with low $\hat{\alpha}$ are expected to consist of ``good" models that are similar to a good quality ensemble. Since these models achieve validation accuracy within a higher range among their group, the number of points they will make mistakes on will be low. This will also result in a low pairwise churn because these models will only have a limited number of test points to disagree on. As observed for validation accuracy, the opposite is again not true for pairwise churn, i.e., models with low pairwise churn will not necessarily have small values of pairwise $\hat{\alpha}$. To summarize, if both the candidate models in a pair are close to the reference function then their pairwise churn will be low.  However, if one candidate model in a pair is close to the reference function, while the other is far away from it, then the pairwise churn between these models can be either high or low. We conclude that \textbf{ smaller values of pairwise $\hat{\alpha}$ does not admit high pairwise churn but low pairwise churn does not imply smaller pairwise $\hat{\alpha}$.}

\begin{figure}[htb]
\centering
\centerline{\includegraphics[width=8.5cm]{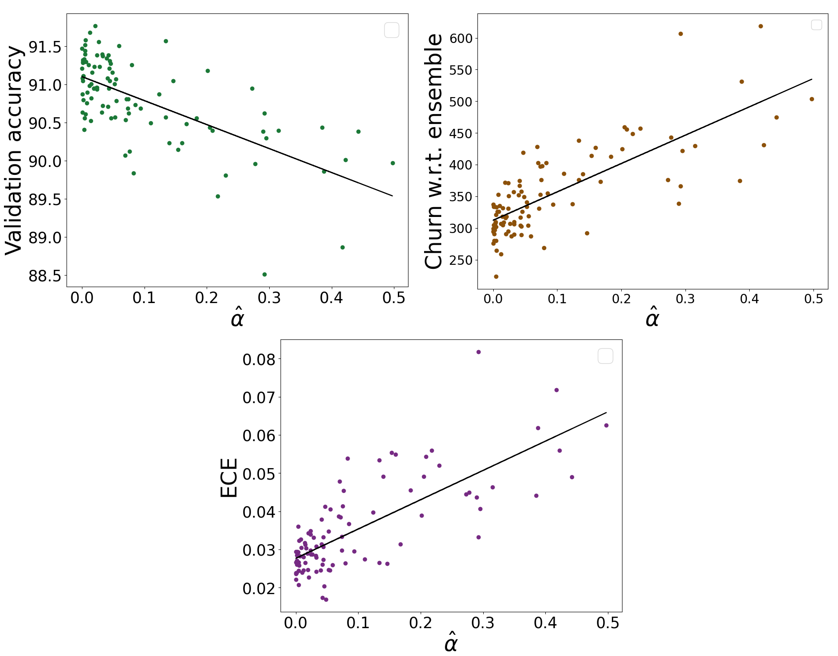}}
 \caption{2-D scatter plot to visualize the relationship of $\hat{\alpha}$ of candidate models in $\randall$  with their average cdf, $\avgcdf$, in terms of (Left) Validation accuracy, (Middle) Churn w.r.t.~their ensemble and (Right) ECE.} 
\label{fig: accecechurn_vs_alpha}
\end{figure}

\subsubsection{Our proposed discrepancy measure is more informative than accuracy}\label{subsection: CNN alphahat informative}
\begin{figure}[htb]
\centering
\centerline{\includegraphics[width=4.5cm]{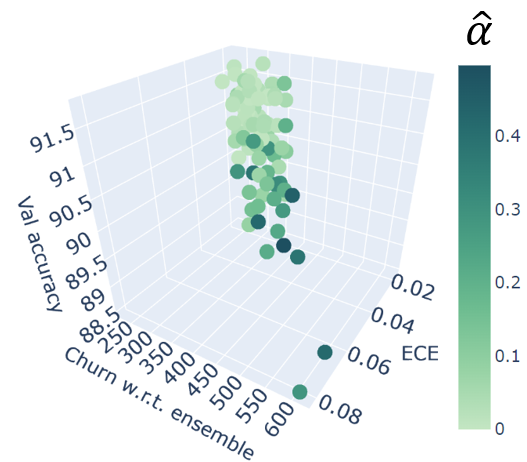}}
\caption{3-D scatter plot to visualize how $\hat{\alpha}$ compares to other metrics for 100 candidate CNN models performing a binary classification task on CIFAR-10.}
\label{fig: 3d_scattercandensemble_batchshuffleinit_alphaep49}
\end{figure}

\begin{table}[tbh]
\caption{Table showing $\hat{\alpha}$ values for CNN models with similar validation accuracy and their corresponding values achieved in other metrics.}
\label{tab:CNN stats}
\centering
\resizebox{\columnwidth}{!}{\begin{tabular}{|c|c|c|c|c|c|}
\hline
&\textbf{ECE} & \textbf{Churn w.r.t.~} & \textbf{Validation} & \textbf{Average} & $\mathbf{\hat{\alpha}}$ \\
&& \textbf{ensemble} & \textbf{accuracy} & \textbf{churn} & \\
\hline
 \textbf{Model 1}&0.028 & 297 & 91.108 & 472.15 & 0.002\\ 
 \hline
 \textbf{Model 2}&0.021 & 280 & 91.519 & 458.61 & 0.005\\ 
 \hline
 \textbf{Model 3}&0.026 & 265 & 91.581 & 459.37 & 0.005\\ 
 \hline
 \textbf{Model 4}&0.030 & 309 & 91.157 & 475.91 & 0.016\\ 
 \hline
 \textbf{Model 5}&0.023 & 318 & 91.769 & 476.95 & 0.021 \\
 \hline
 \textbf{Model 6}&0.030 & 371 & 91.382 & 482.45 & 0.024 \\
 \hline
 \textbf{Model 7}&0.029 & 307 & 91.556 & 476.35 & 0.027 \\
 \hline
 \textbf{Model 8}&0.026 & 376 & 91.569 & $\mathbf{513.83 }$ & $\mathbf{0.134 }$\\
 \hline
 \textbf{Model 9}&$\mathbf{0.038 }$ & $\mathbf{425 }$ & 91.182 & $\mathbf{552 }$ & $\mathbf{0.201 }$\\ 
 \hline
\end{tabular}}
\end{table}

We now provide evidence that $\hat{\alpha}$ is more informative than validation accuracy for model comparison. Figure \ref{fig: accecechurn_vs_alpha} and Figure \ref{fig: 3d_scattercandensemble_batchshuffleinit_alphaep49} show the relationship of $\hat{\alpha}$ of candidate models in $\randall$ with three metrics: validation accuracy, churn w.r.t.~the ensemble of candidate models, and their ECE, in 2-D and 3-D respectively. 
If the eCDF of a candidate model is close to the eCDF of the reference function, i.e. for smaller values of $\hat{\alpha}$, the variability in each of the three metrics is confined to a range that includes the best value in each metric, with the worst value not differing too much from the best value. For each metric, we can think of the range of values corresponding to a small $\hat{\alpha}$ as the minimum variability observed in models that are close to their expected distribution. Any model that has achieved a value worse than this range will also have a high $\hat{\alpha}$. Models with low $\hat{\alpha}$ are within a higher accuracy range among their group and have ECE and churn w.r.t.~ensemble within a low range. However, the converse is not true, i.e. high accuracy or low churn w.r.t.~ensemble or low ECE alone does not imply low $\hat{\alpha}$. These models can achieve a value that is within a good range of values in one metric but can perform poorly in another metric, and this is reflected in the corresponding $\hat{\alpha}$ value. To see this, we focus on models that have achieved a similar high validation accuracy and look at other metrics like ECE, average churn for each model w.r.t.~ all other models, and churn w.r.t.~ their ensemble in Table \ref{tab:CNN stats}. $\hat{\alpha}$ is low if all three metrics fall within the previously discussed good range of values. An increase in $\hat{\alpha}$ (denoted in bold) in Table \ref{tab:CNN stats} indicates a reduction in quality in one or more of these metrics. We can see that most models with similar validation accuracy in Table \ref{tab:CNN stats} have needed only a small trimming level ($\hat{\alpha} \le 0.05$) to not reject the null. But for model 8 and model 9, we see that despite having similar validation accuracy to other models in the group, $\hat{\alpha}$ values are large due to a reduction in quality in other metrics (denoted in bold).

\subsection{Application in Transfer Learning}\label{subsec: Transfer Learning}
\begin{figure}[thb]
\begin{minipage}[b]{0.49\linewidth}
\centering
  \centerline{\includegraphics[width=4cm]{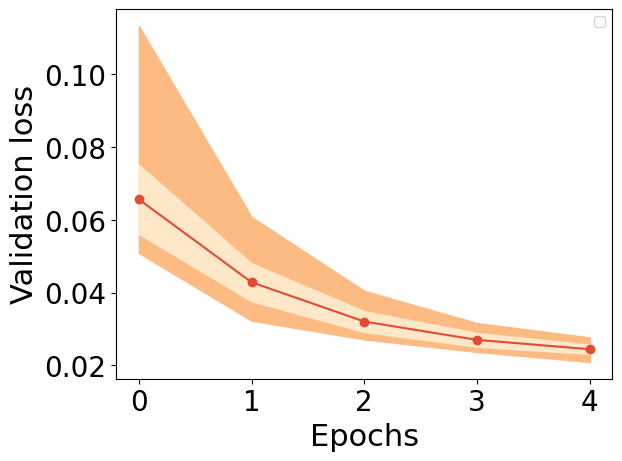}}
\end{minipage}
\begin{minipage}[b]{0.49\linewidth}
\centering
  \centerline{\includegraphics[width=4cm]{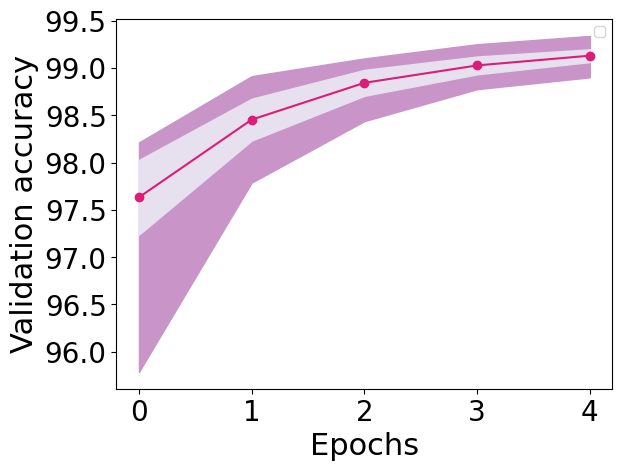}}
\end{minipage}
\caption{ (Left) A plot showing the evolution of validation loss of pre-trained ViT models, over different epochs. The solid red dots represent the mean of validation loss over 45 seeds at each epoch, the light-colored area represents one standard deviation, and the orange area represents the maximum and minimum values at that epoch. (Right) A plot showing the evolution of validation accuracy of pre-trained ViT models, over different epochs. The solid red dots represent the mean of validation accuracy over 45 seeds at each epoch, the light-colored area represents one standard deviation, and the purple area represents the maximum and minimum values at that epoch.}
\label{fig: training stats ViT}
\end{figure}

We conclude our experiments by demonstrating the usefulness of our framework in transfer learning. A common transfer learning setup is when a model trained on one task (usually on a large dataset) is fine-tuned to perform a different, often related task, using a smaller dataset. The smaller dataset (target domain) is assumed to be from the same distribution as the larger dataset (source domain). The pre-trained models in transfer learning are typically very large, designed to capture a wide range of useful features from a vast amount of data, allowing them to generalize well to a variety of tasks.

We use a Vision Transformer(ViT) variant~\cite{liu2021swin}, pre-trained on ImageNet, to perform a downstream binary classification task on CIFAR-10 (as described in Section \ref{sec: experiments}). The pre-trained model is available as part of the Hugging Face Community~\cite{wolf2019huggingface}. We only train the final task-specific classification layer for 5 epochs (till close to convergence as visualized in Figure \ref{fig: training stats ViT}) and rely on already fine-tuned hyperparameters and the pre-trained weights of the upstream task to ``fine-tune" the downstream binary classification task over random seeds only. Thus, keeping the fine-tuning regime and pre-trained weights fixed, we only vary the random seed that controls different sources of randomness in the training procedure (initialization and batch order during SGD), generate 90 models, and observe the variability over random seeds for this fixed setup. As observed by Picard et al.~\cite{picard2021torch}, there is less variability in validation accuracy over random seeds when using pre-trained models. 

\begin{figure}[tbh]
 \centering
 \begin{minipage}[b]{0.49\linewidth}
\centering
  \centerline{\includegraphics[width=4.3cm]{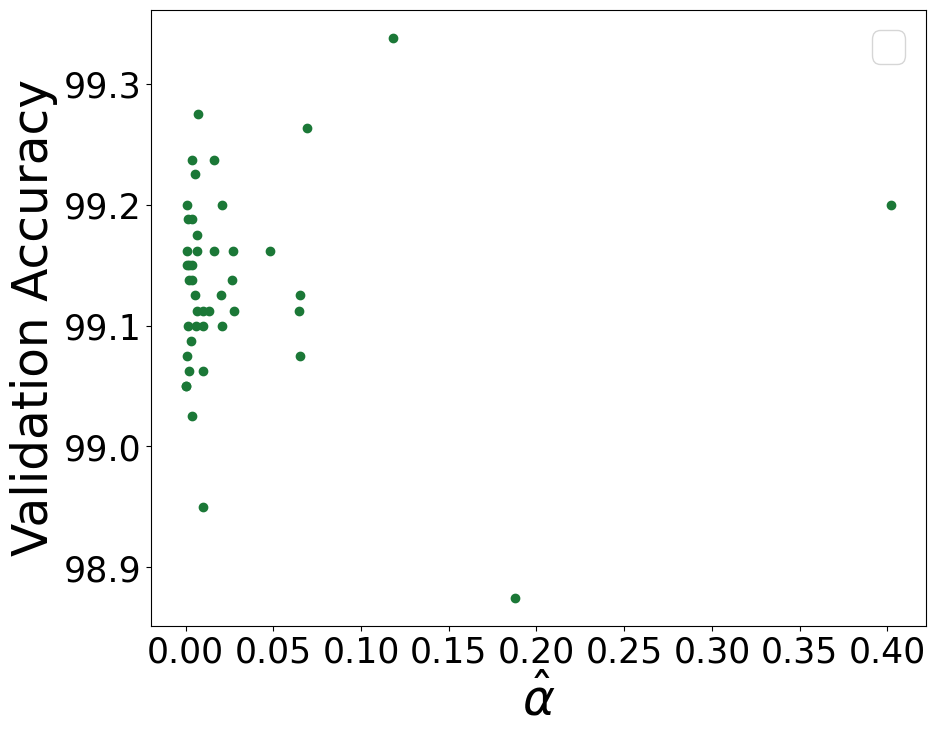}}
\end{minipage}
\begin{minipage}[b]{0.49\linewidth}
\centering
  \centerline{\includegraphics[width=4.3cm]{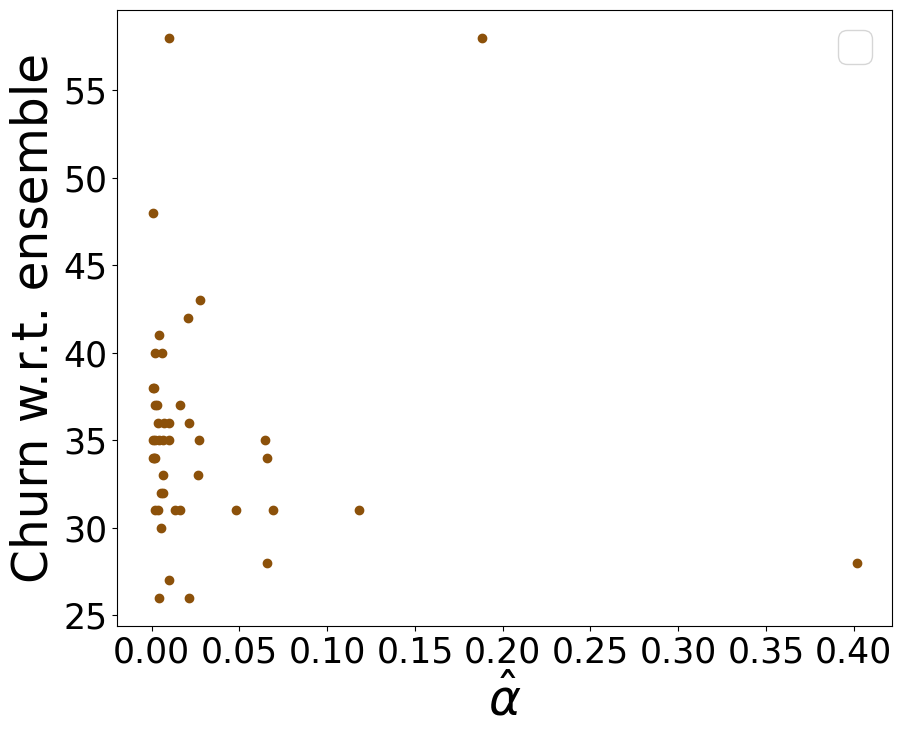}}
\end{minipage}
\begin{minipage}[b]{0.34\linewidth}
\centering
  \centerline{\includegraphics[width=4.3cm]{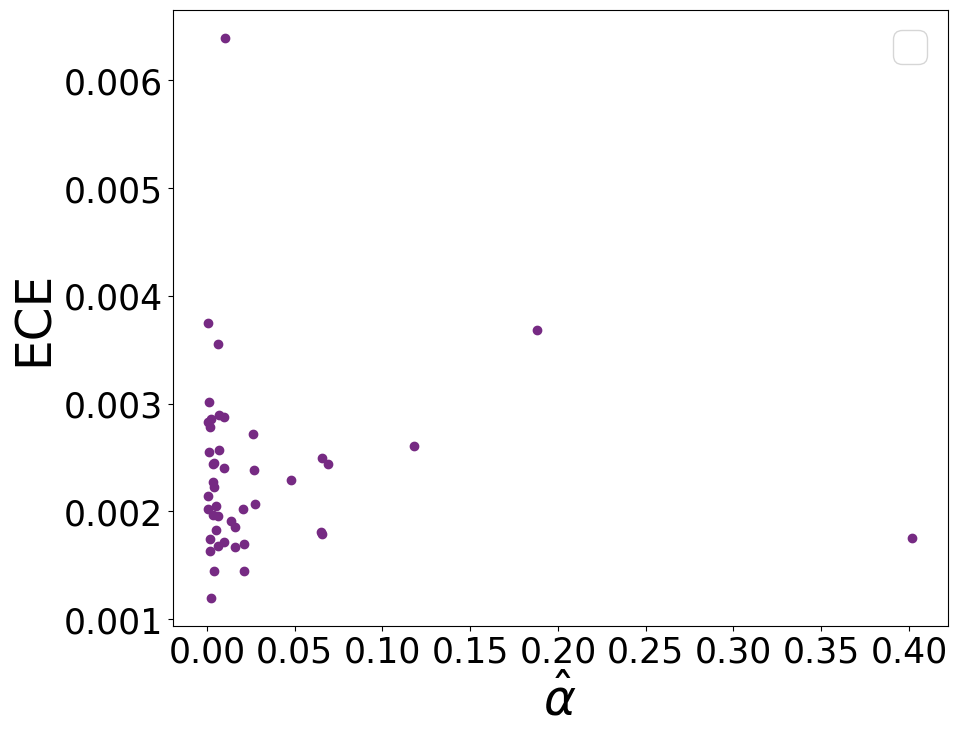}}
\end{minipage}
\caption{2-D Scatter plot to visualize how $\hat{\alpha}$, computed using the robust KS-test against the reference $\avgcdf$, relates to validation accuracy for 45 candidate ViT models (each dot in the plots) differing in random seeds.}\label{fig:alphahat_pretrained} 
\end{figure}

We create our reference function $\avgcdf$ using the first 45 models and treat the remaining 45 models as our candidate models $\multcandcdf$, $l \in \Ncandmodels$, for comparison with the reference. We conduct the same experiment as in Section \ref{subsection: CNN alphahat informative}, where we run our robust hypothesis test between the reference function $\avgcdf$ and candidate models $\multcandcdf$ to compute our proposed discrepancy measure $\hat{\alpha}$. Following similar reasoning to Section \ref{subsection: CNN alphahat informative}, in Figure \ref{fig:alphahat_pretrained}, we observe how our proposed metric is more informative than validation accuracy. The measure $\hat{\alpha}$ is indicative of the quality of models in terms of other metrics like ECE, churn w.r.t.~an ensemble, and average churn of candidate models w.r.t.~all other models in the pool alongside accuracy. A higher $\hat{\alpha}$ is usually accompanied by a reduction in quality in one or more of these three metrics, denoted in bold in Table \ref{tab:vit table}. Although pre-trained large-scale models provide us with very high validation accuracy and less variability in different metrics, our experiment demonstrates how smaller values of $\hat{\alpha}$ are still helpful in identifying the best representatives of the training procedure.

\begin{table}[tbh]
\caption{Table showing $\hat{\alpha}$ values for pre-trained ViT models with similar validation accuracy and their corresponding values achieved in other metrics.
\label{tab:vit table}}
\resizebox{\columnwidth}{!}{\begin{tabular}{|c|c|c|c|c|c|}
\hline
&\textbf{Average} & \textbf{Churn w.r.t.~} & \textbf{Validation} & \textbf{ECE} & $\mathbf{\hat{\alpha}}$\\
&\textbf{churn}& \textbf{ensemble}& \textbf{ accuracy}& &\\
\hline
\textbf{Model 1} &44.82 & 26 & 99.2  & 0.002 & 0.004\\
 \hline
 \textbf{Model 2} &45.88 & 24 & 99.2  & 0.001 & 0.007\\
 \hline
 \textbf{Model 3} &48.33 & 30 & 99.3 & 0.002  & 0.026 \\ 
 \hline
 \textbf{Model 4} &47.97 & $\mathbf{34}$ & 99.2  & 0.001 & $\mathbf{0.069}$\\
 \hline
 \textbf{Model 5} &$\mathbf{49.44}$ & 28 & 99.25 & $\mathbf{0.003}$ & $\mathbf{0.118}$\\
 \hline
\end{tabular}}
\end{table}

\begin{figure}[hbt]
 \begin{minipage}[b]{0.49\linewidth}
\centering
  \centerline{\includegraphics[width=4.4cm]{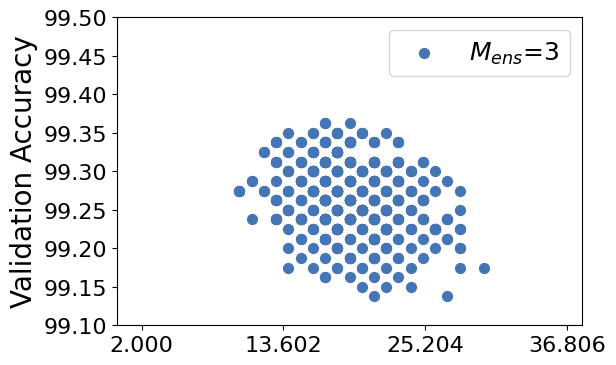}}
\end{minipage}
\begin{minipage}[b]{0.49\linewidth}
\centering
  \centerline{\includegraphics[width=4.16cm]{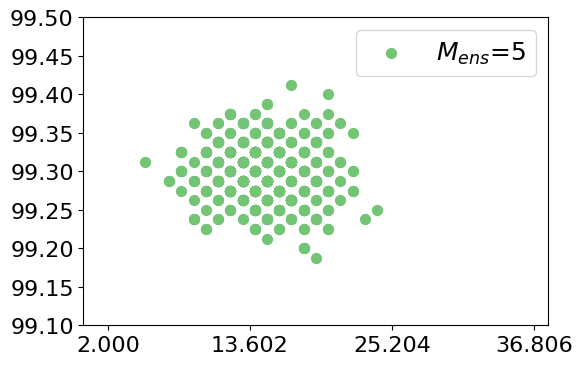}}
\end{minipage}
\centering
 \begin{minipage}[b]{0.49\linewidth}
\centering
  \centerline{\includegraphics[width=4.4cm]{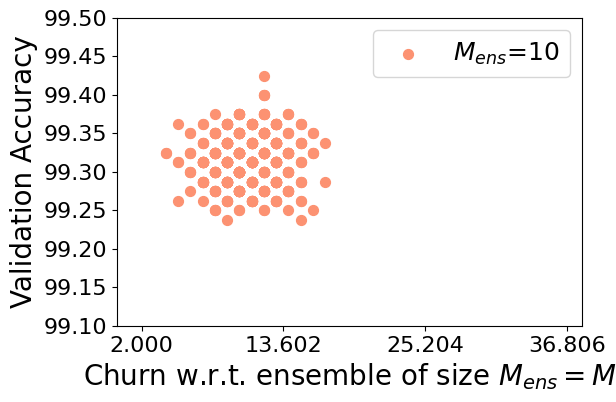}}
\end{minipage}
\begin{minipage}[b]{0.49\linewidth}
\centering
  \centerline{\includegraphics[width=4.16cm]{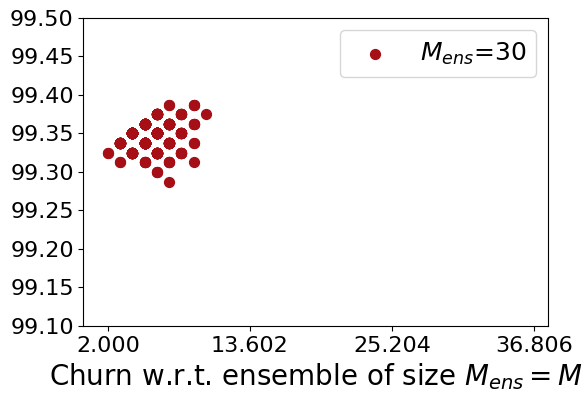}}
\end{minipage}
\caption{2-D scatter plot to visualize the relationship among ViT ensemble models in terms of validation accuracy, and churn w.r.t.~an ensemble of size $\Nensemble=\Nmodels$. For each value of $\Nensemble$, we choose $\Nensemble$ candidate models to form one ensemble model and repeat this experiment 500 times through sampling with replacement to create 500 ensemble models. Thus, each dot with a fixed color represents one out of these 500 ensemble models and each color indicates the value of $\Nensemble$ or the number of candidate models in the pool. Several ensemble models have achieved the same accuracy or churn resulting in fewer than 500 visible models.}\label{fig: Vit Testacc churn ensembles}
\end{figure}

\begin{figure}[tbh]
 \begin{minipage}[b]{0.49\linewidth}
\centering
  \centerline{\includegraphics[width=4.4cm]{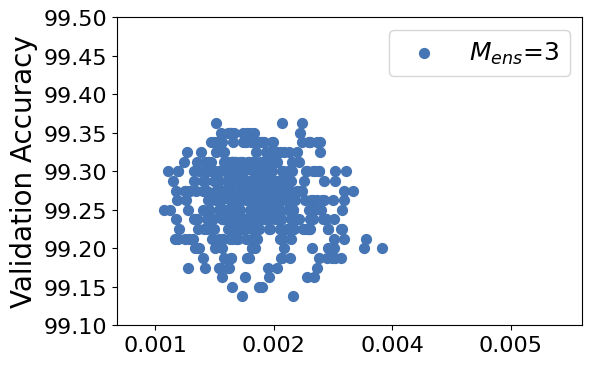}}
\end{minipage}
\begin{minipage}[b]{0.49\linewidth}
\centering
  \centerline{\includegraphics[width=4.16cm]{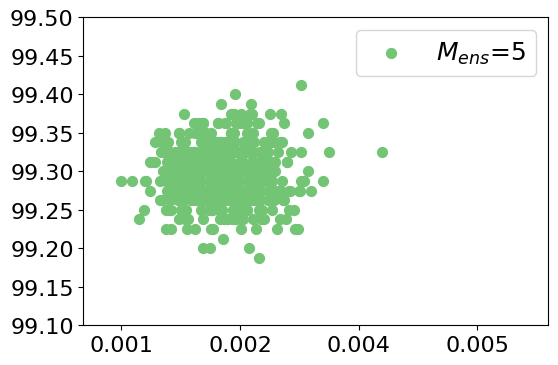}}
\end{minipage}
\centering
 \begin{minipage}[b]{0.49\linewidth}
\centering
  \centerline{\includegraphics[width=4.4cm]{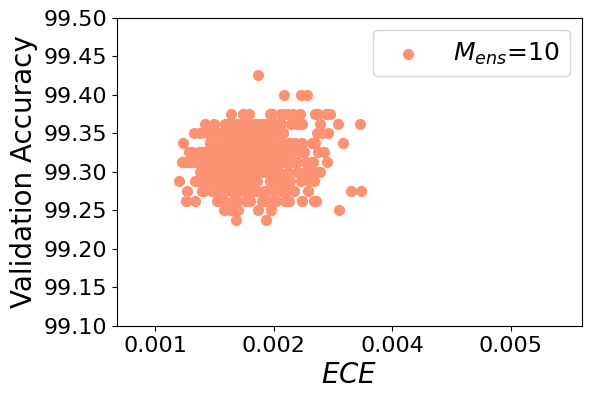}}
\end{minipage}
\begin{minipage}[b]{0.49\linewidth}
\centering
  \centerline{\includegraphics[width=4.16cm]{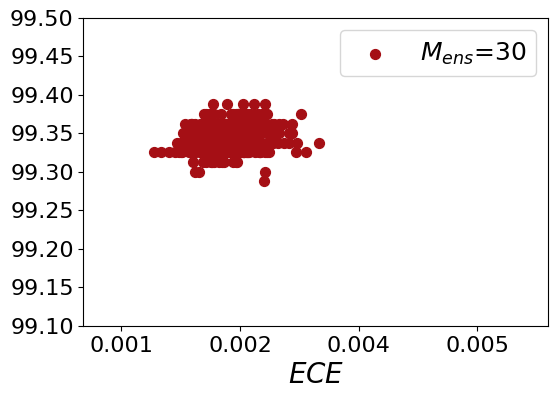}}
\end{minipage}
 \caption{2-D scatter plot to visualize the relationship among ViT ensemble models in terms of validation accuracy, and ECE. We follow the experiment in Figure \ref{fig: Vit Testacc churn ensembles} to create the ensemble models.}\label{fig: Vit Testacc ece ensembles}
\end{figure}
For the next set of experiements, we use the first $\Nmodels=45$ models to create a reference function $\avgcdf$ and set the rest $\Ncandmodels = 45$ models to create the eCDF of a deep ensemble $\cdfavg$. We choose different ensemble sizes, $\Nensemble \in [3, 5, 10, 30]$, to observe the variability of ensembles in different metrics. Figure \ref{fig: Vit Testacc churn ensembles}, and Figure \ref{fig: Vit Testacc ece ensembles} show the relationship among different metrics for ensembles of the pre-trained candidate models. Since all the individually trained models achieved very high validation accuracy and low variability, the ensembles did not result in significant performance gains over their constituents. 
This low variability also resulted in a lower $\hat{\alpha}$ for most pre-trained candidate models, as seen in Figure \ref{fig:alphahat_pretrained}. If all models performed similarly, ensembling them would not make much difference. Thus, moderate variability in model performance may be required for the pool of candidate models to benefit from ensembling, which implies that $\hat{\alpha}$ for all candidate models cannot be too low.

The low variability observed in pre-trained ViT models in the above setting does not generalize to all types of pre-trained models. Dodge et al. demonstrated that finetuning pre-trained language models, like BERT, over weight initialization, and data ordering, results in significant improvement in performance metrics, indicating the instability of the training process in large language models~\cite{dodge2020finetuning}. They provide evidence that some data
orders and initializations are better than others and emphasize the need for more rigorous reporting of benchmark model performance for comparisons across different architectures. However, like most previous works, they also use accuracy as their metric to analyze fine-tuning variability. Our metric $\hat{\alpha}$ can add an extra layer of reliability by identifying ``good'' seeds that closely represent the expected training behavior.

 Through these experiments, we highlight the importance of treating the seeds that control random elements in a DNN as a separate hyperparameter that needs ``tuning" and propose a framework to reliably select seeds that do not rely solely on commonly used performance metrics like validation accuracy. When good seed selection is needed to account for the variability caused by random seeds, we recommend using a rule of thumb such as exploring at least 30 seeds and choosing the seed that generates a model with high validation accuracy and small $\hat{\alpha}$.

\section{Conclusion and Future Work}
In this work, we propose a framework for random seed selection of a DNN with a fixed architecture and a fixed hyperparameter setting. Our proposed framework is based on a robust two-sample hypothesis testing problem that uses trimming of the eCDF of samples obtained from the logit gap function. Our test assesses the closeness of candidate models in a pool with their
 expected eCDF by down-weighting that
 part of the data that has a greater influence on the dissimilarity. We also provide some evidence that our new measure, the trimming level
 $\hat{\alpha}$, could be a more informative metric to assess model performance than commonly used test/validation accuracy. This allows us to perform random seed selection in a more principled fashion instead of relying solely on trial-and-error methods or metrics like validation accuracy. Although our paper focused on the influence of the random seed on model variability, our method can be extended to investigate the influence of any hyperparameter. 
 
 Since the usefulness of our methodology relies on ensembles achieving significant performance gains over their constituents, we investigate the behavior of the ensemble as the number of models in the ensemble pool increases. Our proposed metric $\hat{\alpha}$ can also be used as a stopping criterion for the number of different seeds that need to be explored for an ensemble model to reliably approximate the expected eCDF of the logit gap distribution of candidate models, and to have less variability in different performance metrics. Although ensembling by model averaging in the output space is quite straightforward and usually results in performance gains, this comes at the cost of reduced interpretability, since such ensembles do not learn any parameters or features. By selecting models that are close to their ensembles in the output space of a DNN, instead of the ensemble itself, our framework leaves room for model interpretability while at the same time maintaining high accuracy compared to their counterparts. Our methodology is useful in critical application areas like credit risk assessment where a single model is often preferred over the improved accuracy of an ensemble due to interpretability concerns~\cite{florez2015enhancing}.

There is a common belief in the ensemble learning literature that increasing  diversity among ensemble members will improve the quality of the ensembles~\cite{abe2023pathologies}~\cite{theisen2023ensembles}. If all models were to perform exactly the same then the ensemble of these models are not going to perform any better than a single model in the ensemble. As observed from the experimental results in the transfer learning application, low diversity in models would indicate lower values of $\hat{\alpha}$ for all candidate models. However, imposing too much diversity among ensemble members can also be detrimental to the ensembles performance as increasing diversity can sacrifice overall model performance. In this scenario, we would likely find many models with large $\hat{\alpha}$. Through our test, $\hat{\alpha}$ will select models that are closer to a poor performing ensemble, which may lead to selecting random seeds that achieve poor accuracy. Thus, in scenarios where ensembling by model averaging might not lead to significant performance gains over candidate models in the pool or hurt performance, future extensions of this work include investigating when candidate models form effective ensembles.

We are also interested in extensions to multi-class
classification and exploring robust two-sample hypothesis testing based on other distance metrics like the Wasserstein metric~\cite{AlvarezEsteban2012}. 
In this work, we focused on samples from the logit gap function as our probe to understand deep net variability. We can look at other functions of the trained models, such as the Jacobian or the Neural Tangent Kernel of functions learned by these models~\cite{jacot2020neural}. The eigen-distribution of these matrices for instance can inform us how well individual candidate models can generalize to test data.

\begin{appendices}
\section{Proof of Theorem \ref{Main theorem}}

\label{sec: Analysis}

\begin{IEEEproof}
Both $\condcdf$ and $\avgcdf$ take averages over $\Nmodels$ parameters. Define     \begin{align}
    A_{k}(t) &\doteq \E_{\datadist |\param_{k}}\bracks*{ 
        \Indic{{m(\mbf{x},\param_{k}})\leq t} }, \textrm{ and}\\
    B_{k}(t) &\doteq \frac{1}{ N}\sum_{j=1}^{ N} 
        \Indic{m(x_{j}; \param_{k}) \leq t }
    \end{align}
Then we can upper bound $\normifty{ \condcdf - \avgcdf }$ by
    \begin{align}
    \normifty{ \condcdf - \avgcdf } \le \frac{1}{\Nmodels} \sum_{k=1}^{\Nmodels} 
        \normifty{ A_k - B_k }.
    \label{eq:cdf:trianglebound}
    \end{align}
For each $\param_k$, $B_k$ is the eCDF of samples drawn from a distribution with CDF $A_k$. By the DKW inequality~\cite{dvoretzky1956asymptotic},
    \begin{align}
    \P_{\datadist|\param_k}\parens*{ \normifty{ A_k - B_k } > \delta_b }
    \le 
    2 \exp(-2 N\delta_{b}^{2}).
    \end{align}
Taking a union bound over $\dparam = \{\param_k : k \in [\Nmodels]\}$ we get,
    \begin{align}
    \P_{\datadist|\dparam}\parens*{ \exists k \in [\Nmodels] \text{\ s.t.\ } \normifty{ A_k- B_k } > \delta_b }\notag\\
    \le 
    2 \Nmodels \exp(-2 N\delta_{b}^{2}).
    \end{align}
Therefore, each term on the right-hand side of \eqref{eq:cdf:trianglebound} is smaller than $\delta_b$ w.h.p.~and as desired,
    \begin{align}
    \P_{\datadist|\dparam}\parens*{ 
    \normifty{ \condcdf - \avgcdf } \le \delta_b 
    }\notag\\
    \ge 1 - 2 \Nmodels \exp(-2 N\delta_{b}^{2}).
    \end{align}
\end{IEEEproof}

\section{More details on trimming}

\subsection{Impartial trimming}\label{subsec: Impartial trimming}
In Section \ref{sec: Robust KS test}, we introduce a robust version of the KS-test, where we allow outliers in samples from the null hypothesis using $\alpha$-trimming of distributions. For completeness, we present the concepts on $\alpha$-trimming of a distribution w.r.t.~a reference distribution as introduced by Alvarez-Esteban et al.~\cite{alvarez2008trimmed }. Trimming allows us to assume some outliers in samples that did come from the null distribution and helps quantify the fraction of these outliers. 

Let $k$ be the number of trimmed observations, and let $\alpha$ be the trimming fraction, which implies $k \le n\alpha$. Given n i.i.d samples $\{x_{i}\}_{i=1}^{n}$ with probability distribution $P_{1}$, the empirical measure can be defined as $\frac{1}{n}\sum_{i=1}^{n}\delta^{\text{Dirac}}({x_{i}})$, where $\delta^{\text{Dirac}}({x})$ is the Dirac measure. To not reject the null, we can remove outliers by assigning a weight of 0 to the bad observations in the sample and adjusting the weight of good observations by $\frac{1}{n-k}$.  However, we may not want to completely get rid of samples from the feasible set and instead downplay the importance of bad observations by modifying the empirical measure to be $\frac{1}{n}\sum_{i=1}^{n}w_{i}\delta_{x_{i}}$, where $0 \leq w_{i} \leq \frac{1}{(1 - \alpha)}$, and $\frac{1}{n}\sum_{i=1}^{n} w_{i} = 1$. This is called an \textit{impartial trimming} of the probability measure $P_{1}$.
For the remainder, we use the term impartial trimming interchangeably with trimming. 

An $\alpha$-trimming of the distribution $P_{1}$, denoted by $R_{\alpha}(P_{1})$, is defined as~\cite[Definition 1]{alvarez2008trimmed},
\begin{equation}
\resizebox{225pt}{!}{%
$
\label{eq:11}
    \mathcal{R}_{\alpha}(P_1) =  \{ P \in \mathcal{P}: P\ll P_{1},\frac{dP}{dP_1} \leq \frac{1}{(1 - \alpha)} P_1 \text{ a.s}. \}
$%
}
\end{equation}
where $P_{1},P$ are probability measures on $\mathbb{R}$, $0 \leq \alpha \leq 1$, and we say $P$ is an $\alpha$-trimming of $P_{1}$ if $P$ is absolutely continuous w.r.t.~$P_{1}$ and satisfies the above definition.
The set of $R_{\alpha}(P_{1})$, the $\alpha$-trimmings of $P_{1}$, can be characterized in terms of the trimming function $h$. The function $h$ determines which zones in the distribution $P$ are downplayed or removed. 

Let $\mathcal{C}_{\alpha}$ be the class of absolutely continuous functions $h:[0,1] \rightarrow [0,1]$, such that $h(0)=0$, and $h(1)=1$, with derivative $h'$, such that $0 \leq h'\leq \frac{1}{1 - \alpha}$. For any real probability measure $P_1$, the following holds~\cite[Proposition 1 a.]{alvarez2008trimmed},
\begin{equation}
\resizebox{225pt}{!}{%
$
    \mathcal{R}_{\alpha}(P_1) = \{P \in \mathcal{P} : P(-\infty,t]=h(P_1(-\infty,t]),h \in C_{\alpha} \}. 
$%
}
\end{equation}
\subsection{Trimmed Kolmogorov-Smirnov distance}\label{subsec: Trimmed Kolmogorov-Smirnov distance}
Given two distribution functions, $\expectedcdf$ and $\candtruecdf$, as defined in Section \ref{subsec: true cdfs}, the KS distance is given by,
\begin{equation}
    d(\expectedcdf,\candtruecdf) = \sup_{x \in \mathbb{R}} \lvert \expectedcdf(x) - \candtruecdf(x) \rvert.
\end{equation}
We similarly define the $\alpha$-trimmed KS distance functional as,
\begin{equation}
    d(\expectedcdf, R_{\alpha}(\candtruecdf)) = \min_{\tilde{F} \in \mathcal{R}_{\alpha}(\candtruecdf)} d(\expectedcdf, \tilde{F}).
\end{equation}
The plug-in estimator for $d(\expectedcdf, R_{\alpha}(\candtruecdf))$ is $d(\expectedcdf, R_{\alpha}(\candcdf))$, where $\candcdf$, defined in Section \ref{subsec: empirical cdfs}, is the empirical distribution function based on a sample of $ N$ independent random variables with common distribution function $\candtruecdf$, and $\expectedcdf$ is our null distribution.
A practical computation for $d(\expectedcdf,\mathcal{R}_{\alpha}(\candcdf))$ uses function $\expectedcdf \circ \candtruecdf^{-1}$ to express $d(\expectedcdf, R_{\alpha}(\candtruecdf))$. If $\expectedcdf$ is continuous and $x$ (generated samples) has distribution function $\candtruecdf$, $\expectedcdf \circ \candtruecdf^{-1}$ is the quantile function associated with the random variable $Y=\expectedcdf(x)$. This gives rise to the following lemma and theorem for computing the trimmed KS distance between a theoretical distribution and an eCDF~\cite{del2020approximate},

\begin{lemma}[\cite{barrio2020box}, Lemma 2.4]
If $\candtruecdf,\expectedcdf$ are continuous distribution functions and $\candtruecdf$ is strictly increasing then, 
\begin{align}
    d(\expectedcdf,\mathcal{R}_{\alpha}(\candtruecdf)) &= \min_{h \in C_{\alpha}} \lvert\lvert h - \expectedcdf \circ \candtruecdf^{-1} \rvert\rvert, \hspace{0.5cm} \\
    d(\expectedcdf,\mathcal{R}_{\alpha}(\candcdf)) &= \min_{h \in C_{\alpha}} \lvert\lvert h - \expectedcdf \circ \candcdf^{-1} \rvert\rvert.
\end{align}
\end{lemma}

\begin{theorem}[\cite{barrio2020box}, Theorem 2.5]
Suppose $\Gamma:[0,1] \rightarrow [0,1]$ is a continuous non-decreasing function and let
    \begin{align}
    B(t) &= \Gamma(t) - \frac{t}{1 - \alpha}, \\
    U(t) &= \sup_{t \leq s \leq 1} B(s), \\
    L(t) &= \inf_{0 \leq s \leq t} B(s), \textrm{ and} \\ 
    \tilde{h}_{\alpha}(t) &= \max\bigg(\bigg(\min\bigg(\frac{U(t)+L(t)}{2}\bigg),0\bigg)\frac{-\alpha}{(1 - \alpha)}\bigg).
    \end{align}
Then 
    \begin{align}
    h_{\alpha} \defeq \tilde{h}_{\alpha} + \frac{(\cdot)}{(1 - \alpha)}
    \end{align}   
is an element of $C_{\alpha}$ and $\min_{h \in C_{\alpha}}  \lvert\lvert h - \Gamma \rvert\rvert = \lvert\lvert h_{\alpha} - \Gamma \rvert\rvert=\lvert\lvert \tilde{h}_{\alpha} - \Gamma \rvert\rvert $, with assumptions on $\Gamma$ holding for $\Gamma=\expectedcdf \circ \candtruecdf^{-1}$.
\end{theorem}

In our application we don't have access to $\expectedcdf$, and as discussed in Section \ref{subsec: a new robust two sample test}, our robust hypothesis test is against the reference function $\avgcdf$. Since this function is an average of eCDFs and not a continuous function, for practical computation of $d(\avgcdf,\trimmed_{\alpha}(\candcdf))$, we consider the linearly interpolated CDF of $\avgcdf$, which we denote as $\avgcdfinterp$.
Thus, running the test in  $\eqref{eq: robust2sample_ht}$ is equivalent to running the following test,
\begin{align}
    \max_{z} \min_{\tilde{F} \in \trimmed_{\alpha}(\candcdf)} | \avgcdfinterp(z) - \tilde{F}(z) | \bhtproxy \delta_{a} + \frac{1}{ N},
    \label{eq: robust2sample_htinterp}
\end{align}
where we arrive at the threshold on the right-hand side by considering the inequality $\normifty{\avgcdf-\avgcdfinterp} \le \frac{1}{ N}$ and then applying the triangle inequality.

\section{Additional Experimental Details}

\label{sec: CNN architecture}
\begin{figure}[tbh]
 \centering
   \centerline{\includegraphics[width=7cm]{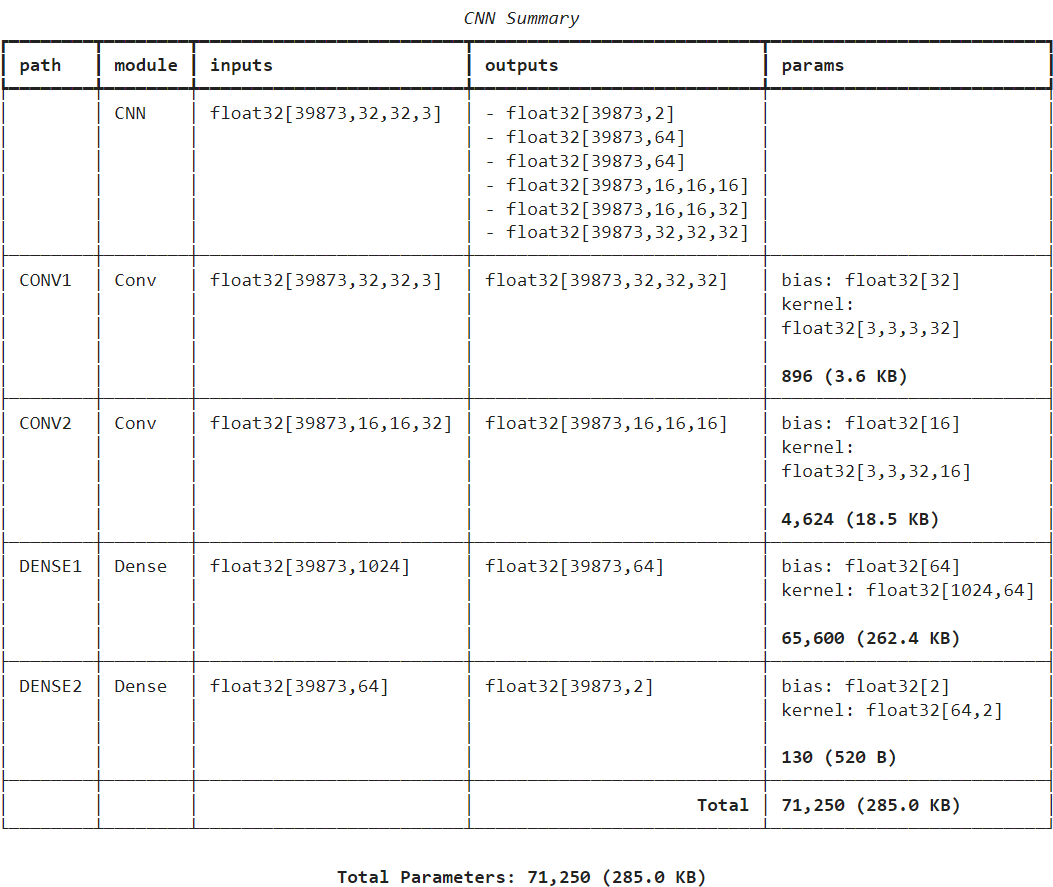}}
 \caption{CNN Architecture}
 \label{fig:architecture}
\end{figure}

Figure \ref{fig:architecture} shows the CNN architecture and parameters used in Section \ref{sec: experiments}.

\end{appendices}

\bibliographystyle{IEEEtran}
\bibliography{IEEEabrv,refs}

\end{document}